\documentclass[preprint,12pt]{elsarticle}

\usepackage{algorithm}
\usepackage{algpseudocode}
\usepackage{bm}
\usepackage{url}
\usepackage{color}
\usepackage{comment}
\usepackage{tabularx}

\newcommand{\argmin}{\mathop{\rm arg~min}\limits}
\newcommand{\argmax}{\mathop{\rm arg~max}\limits}
\newcommand{\bmt}{\bm{\theta}}
\newcommand{\bmT}{\bm{\Theta}}
\newcommand{\bmM}{\bm{M}}
\newcommand{\bmS}{\bm{S}}

\journal{Computational Statistics and Data Analysis}

\begin{document}

\begin{frontmatter}

\title{SMLSOM: The shrinking maximum likelihood self-organizing map}

\author[label1]{Ryosuke~Motegi\corref{cor1}}
\ead{t212d001@gunma-u.ac.jp}

\author[label2]{Yoichi~Seki}
\ead{sekiyoichi@gunma-u.ac.jp}

\cortext[cor1]{Corresponding author}

\affiliation[label1]{
            organization={Graduate School of Science and Technology, Gunma University},
            addressline={1-5-1 Tenjin},
            city={Kiryu},
            postcode={376-8515},
            state={Gunma},
            country={Japan}}

\affiliation[label2]{
            organization={Faculty of Informatics, Gunma University},
            addressline={1-5-1 Tenjin},
            city={Kiryu},
            postcode={376-8515},
            state={Gunma},
            country={Japan}}

\begin{abstract}
Determining the number of clusters in a dataset is a fundamental issue in data clustering. Many methods have been proposed to solve the problem of selecting the number of clusters, considering it to be a problem with regard to model selection.
This paper proposes an efficient algorithm that automatically selects a suitable number of clusters based on a probability distribution model framework. 
The algorithm includes the following two components.
First, a generalization of Kohonen's self-organizing map (SOM) is introduced. In Kohonen's SOM, clusters are modeled as mean vectors. In the generalized SOM, each cluster is modeled as a probabilistic distribution and constructed by samples classified based on the likelihood.
Second, the dynamically updating method of the SOM structure is introduced. In Kohonen's SOM, each cluster is tied to a node of a fixed two-dimensional lattice space and learned using neighborhood relations between nodes based on Euclidean distance. The extended SOM defines a graph with clusters as vertices and neighborhood relations as links and updates the graph structure by cutting weakly-connection and unnecessary vertex deletions. The weakness of a link is measured using the Kullback--Leibler divergence, and the redundancy of a vertex is measured using the minimum description length. Those extensions make it efficient to determine the appropriate number of clusters. Compared with existing methods, the proposed method is computationally efficient and can accurately select the number of clusters.
\end{abstract}


\end{frontmatter}

\section{Introduction}\label{sec:introduction}
Clustering is a fundamental method for the analysis of univariate and
multivariate data. Its applications include data mining, vector quantization, and pattern recognition~\cite{jain2010data}\cite{Du201089}\cite{saxena2017review}.
In the implementation of clustering, the selection of the number of clusters $M$ can be difficult. In most situations in which an application needs to perform clustering, the true number of clusters $M^*$ is usually unknown. The selection of $M$ as $M \ll M^*$ or $M \gg M^*$ could cause misleading results. In many methods that have been proposed to solve the problem of selecting $M$, it is considered to be a problem of model selection~\cite{bouveyron_celeux_murphy_raftery_2019}. Several approaches for model selection have been proposed, including likelihood methods with a penalized term for the model order, and Markov chain Monte Carlo (MCMC) simulation.

In the penalized likelihood method, an optimal $\hat{M}^*$ is selected among candidate models that are obtained by performing clustering with different values of $M$, using a predetermined form of the penalized likelihood.
Several penalized likelihood forms have been proposed from various perspectives.
The Akaike information criterion (AIC)~\cite{akaike1974new} is derived by minimization of the Kullback--Leibler (KL) divergence~\cite{kullback1951information} between the true and estimated models. The Bayesian information criterion (BIC)~\cite{schwarz1978estimating} is derived using Bayesian methodology, and is widely used.
In addition, the minimum description length (MDL)~\cite{rissanen1978modeling} and minimum message length (MML)~\cite{wallace1968information}\cite{wallace1987estimation}, which are derived in terms of coding theory, are also popular.
Another form, integrated classification likelihood (ICL)~\cite{biernacki2000assessing}, is an improvement of BIC for the clustering task.

On the other hand, Bayesian inference methods have also been proposed for model selection. The method of Richardson and Green~\cite{richardson1997bayesian} estimates the posterior probability distribution of $M$ using reversible-jump Markov chain Monte Carlo (RJMCMC)~\cite{green1995reversible}, which is mainly used for density estimation using a Gaussian mixture model (GMM).

The technique of using a model selection criterion to select an optimal model among candidate models is simple; however, it is necessary to carefully determine the initial values of the model parameters. For example, an expectation--maximization (EM) algorithm~\cite{dempster1977maximum} requires the initial value dependence to be considered. Although the method based on Bayesian inference can provide abundant information regarding $M^*$, the MCMC-based method is computationally expensive.

In terms of methods with low initial value dependency and computational efficiency for selecting $M$, approaches using a greedy algorithm have been proposed.
Many of these algorithms search for $M^*$ by splitting each cluster.
The decision as to whether to split a cluster is made using a predetermined split decision criterion.
X-means~\cite{pelleg2000x}, which is a representative method of this approach, uses a K-means algorithm and BIC to apply the split decision criterion.
Some studies on improving X-means have been reported~\cite{hamerly2004learning}\cite{feng2007pg}\cite{kalogeratos2012dip}. 

G-means~\cite{hamerly2004learning} uses a statistical hypothesis test as a split decision criterion, which tests the hypothesis to determine whether the data in a cluster exhibit a Gaussian distribution. If the cluster does not seem to be a single Gaussian, it is split into two clusters. G-means projects the samples within a cluster into one dimension for statistical testing. The projection is the direction of the first principal component of the cluster to be considered splitting.

PG-means~\cite{feng2007pg} does not adopt the splitting method, but it is also an extension of G-means and related to X-means. PG-means assumes the dataset has been generated from a GMM and uses the EM algorithm to estimate the model. PG-means projects the dataset and model (means and covariances) into multiple one-dimensional spaces and tests model fitness in each space. The projections are generated randomly. If any test rejects, the number of clusters is increased by one. 

Dip-means~\cite{kalogeratos2012dip} is another approach that uses a hypothesis test as a split decision method. The unimodality of the cluster is tested using Hartigan's dip test~\cite{hartigan1985dip}, and the cluster is split into two until the distribution within a cluster becomes unimodal. It tests the unimodality of a distance distribution of each sample within a cluster and splits the cluster with the large proportion of rejected samples.

In contrast to a cluster-splitting method such as X-means, methods that remove a cluster that is no longer a good representation of the data distribution also exist.
The method of Figueiredo and Jain~\cite{figueiredo2002unsupervised}, called MML-EM, determines the value of $M$ by fitting a GMM to the data distribution with MML as the objective function.
In their method using the EM algorithm, learning starts from a sufficiently large number of clusters, which are gradually annihilated during the learning process. Clusters that are not supported by most samples during the learning process (i.e., clusters whose mixing probability is close to 0) are removed. The objective function is minimized by the EM algorithm, and cluster annihilation and optimization are performed until no further improvement in the MML occurs. The method has displayed a lower initial value dependence than the standard EM algorithm in artificial and real data experiments.

The above greedy methods use either K-means or the EM algorithm as the learning method. However, K-means is known to be strongly dependent on the initial positions of centroids and to easily converge to a local solution. One of the drawbacks of the EM algorithm is its slow convergence. Although MML-EM can reduce the computational time compared to the standard EM, it remains computationally inefficient compared with methods such as X-means.

Examples of methods with little dependence on the initial positions, yet offering fast convergence, are Kohonen's self-organizing maps (SOMs)~\cite{kohonen1982self} and neural gas (NG)~\cite{martinetz1993neural}, which are classical vector quantization methods. These methods have advantages and disadvantages.
NG is less likely to converge to a local solution than SOM, but requires more computational time for learning than SOM.

Our objective is to construct a fast algorithm with a low initial dependence for selecting a suitable number of clusters. As with MML-EM, the algorithm starts with a sufficiently large $M$ and searches for a suitable number of clusters while decreasing $M$.
Although SOM and NG are popular for clustering tasks, the $O(M\log M)$ computations of NG learning compared with the $O(M)$ of SOM are unacceptable for our purpose. Therefore, we select SOM as the learning method and propose a greedy method for automatically selecting $M^*$ based on the SOM learning rule. 

The proposed method is the decreasing approach similar to MML-EM. The method repeats learning by SOM and removing an unnecessary cluster based on the MDL criterion. Each cluster is modeled as a probabilistic distribution and constructed by samples based on maximum likelihood classification. So we call the method the shrinking maximum likelihood self-organizing map (SMLSOM).

The following two extensions to Kohonen's SOM are made for our approach. First, in Kohonen's SOM, clusters are constructed as sample averages, but our method constructs each cluster as a probabilistic model. Therefore, we extended the SOM learning to a probabilistic setting.
Second, in Kohonen's SOM, each cluster is tied to a node of a two-dimensional lattice map, and the map structure is fixed. However, our method removes an unnecessary cluster in the learning; the map structure should vary. Therefore, we introduced a dynamically update method of the map structure to decrease the number of clusters. This update method is made possible by using a graph structure as the map in combination with two procedures: weakly-connected link cutting and unnecessary node deletion.

The remainder of this paper is organized as follows. In Section~\ref{sec:background}, we present the work related to the proposed method. In Section~\ref{sec:algorithm}, we discuss the SMLSOM algorithm, which comprises the SOM based on the maximum likelihood method and the updated map structure. Section~\ref{sec:experiment} reports the experimental results obtained using artificial and real data. Section~\ref{disccusion} discusses the strengths and weaknesses of the proposed method. Section~\ref{sec:conclusion} concludes the paper.
\section{Background}
\label{sec:background}
\subsection{Self-organizing map}
An SOM~\cite{kohonen1982self}\cite{kohonen} is a learning model based on the concept of the structure of the human visual cortex, and offers a method for projecting high-dimensional data onto a low-dimensional lattice space. The lattice space of the SOM represents the topological structure of the input space discretized with $M$ reference vectors, and it is referred to as a map. A two-dimensional lattice, such as a square or hexagonal lattice, is typically used for the structure of the map. SOMs are widely used for vector quantization~\cite{gray1984vector}, clustering~\cite{vesanto2000clustering} and data visualization.

The learning algorithm of the SOM is divided into two stages. First, the Euclidean distance between the input sample $\bm{x}\in \mathcal{R}^p$ and the reference vector of each node $\bm{\mu}_m\in \mathcal{R}^p$, which is associated with the input space, is calculated, and the {\it winner} node $c$ with the smallest distance is determined. Second, the reference vector of each node is updated such that it closely approximates the input.
\begin{enumerate}
 \item Find the winner
       \begin{equation}
    c = \argmin _m \|\bm{x}-\bm{\mu}_m\|, \quad m=1,\,2,\,\ldots ,\,M.
       \end{equation}
 \item Update nodes at iteration $\tau$
       \begin{equation}
    \begin{array}{l}
     \bm{\mu}_m = \bm{\mu}_m + h_{c\,m}(\tau)\Delta_\tau \bm{\mu}_m \\
     \qquad {\rm where}\ \Delta_\tau \bm{\mu}_m = \alpha (\tau) [\bm{x}-\bm{\mu}_m],
    \end{array}
    \label{eq:som update rule}
       \end{equation}
\end{enumerate}
where $\alpha (\tau)$ is the learning rate ($0 < \alpha (\tau) < 1$) that controls the degree of learning, and $h_{c\,m}(\tau)$ is the neighborhood function that adjusts the degree of learning according to the distance on the map between nodes $c$ and $m$. These are monotonically decreasing scalar functions with respect to the number of learning iterations $\tau$. The learning iterates until the maximum iterations $\tau_{\rm max}$.

Note that the Gaussian kernel function is often used for the neighborhood function as follows:
\begin{equation}
 h_{c\,m}(\tau) = \exp\left[{-\frac{d(c,\,m)}{2\sigma ^2(\tau)}}\right],
\end{equation}
where $d(c, \,m)$ is the Euclidean distance on the map between nodes $c$ and $m$, and $ \sigma (\tau)$ is a monotonically decreasing scalar function with respect to $\tau$ and controls the degree of ``nearness.'' Also, the following simple functions are often used:
\begin{equation}
 h_{c\,m}(\tau) = \left\{\begin{array}{cc}
                   1 & d(c,\,m) \leq r(\tau)\\
                          0 & d(c,\, m) > r(\tau)
                         \end{array},\right.
 \label{eq:neighborhood function_2}
\end{equation}
where $r(\tau)$ is a monotonically decreasing scalar function with respect to $\tau$ and called in ``neighborhood radius.''

Unlike K-means, the learning process of an SOM entails ``soft-to-hard'' learning. In the K-means algorithm, the input $\bm{x}$ only updates its nearest node. Therefore, the input and node have a one-to-one correspondence. In contrast, in the SOM, the input and the nodes are in a one-to-many relationship, as defined by the neighbor function $h_{cm}$ in the earlier stage of learning. They eventually converge to attain one-to-one correspondence, increasingly resembling the K-means algorithm as learning progresses. By controlling the learning process in this way, in the ``soft'' learning phase, the nodes gather around the center of the region where the density is high. Then in the ``hard'' learning phase, each node moves to the centroid of the area it represents. The ``soft-to-hard'' learning used by the SOM means that it is expected to be less likely to converge to local minima than K-means.

\subsection{Minimum description length criterion}
\label{sec:mdl}
The MDL~\cite{rissanen1978modeling}\cite{hansen2001model} is a model selection criterion according to which the best model is the one that can encode the given data in the most concise manner.

Assume that each sample of a dataset $\bm{X} = (\bm{x}_1\,\bm{x}_2\,\cdots\,\bm{x}_n)^t$ follows the probability distribution $f$ independently.
\begin{equation}
  f(\bm{x}\mid \bm{\Psi}_M)=\sum _{m=1}^M \pi _m f(\bm{x}\mid \bmt _m),
  \label{183800_19Jan17}
\end{equation}
where $\pi_1,\,\pi _2,\,\ldots,\,\pi _M$ are the {\it mixing
  probabilities} that satisfy $\pi_m \geq 0$ and $\sum _{m=1}^M \pi_m = 1$, $\bmt_m$ are the parameters of the $m$th component, and $\bm{\Psi}_M = \{\bmt
_1,\,\ldots,\,\bmt _M,\,\pi _1,\,\ldots ,\,\pi_M\}$ are the parameters necessary to specify the mixture.

The likelihood of the model is denoted as follows.
\begin{equation}
  L(\bm{\Psi}_M)=\prod_{i=1}^n \sum _{m=1}^M \pi _m f(\bm{x}_i\mid \bmt _m).
  \label{likelihood_mix}
\end{equation}

According to the information theory, the code length of $\bm{X}$ encoded by $f$ is proportion to the log-likelihood. Let $\hat{\bm{\Psi}}$ be the maximum likelihood estimator; the MDL criterion selects the model that minimizes the following code length:
\begin{equation}
 \label{mdl_mixmodel}
 -\log L(\hat{\bm{\Psi}}_M) + \frac{{\rm df}(\hat{\bm{\Psi}}_M)}{2} \log
 n,
\end{equation}
where ${\rm df}(\hat{\bm{\Psi}}_M)$ is the degree of freedom of the model. In Eq.~(\ref{mdl_mixmodel}), the first term is the code length of $\bm{X}$, and the second term is the code length of the model itself.

When Eq.~(\ref{mdl_mixmodel}) is multiplied by 2, it coincides with the BIC.

\section{Proposed Method}
\label{sec:algorithm}
Our method uses a map that comprises nodes and the links between nodes. The node represents the parameters of the probability distribution model $\bmt _m$, and the link represents the two linked nodes as neighbors. This neighborhood relationship is important for SOM learning rules.

The algorithm has two components. The first is a ``soft-to-hard'' learning step that takes $M$ probability distribution models and the map as input, and learns the model parameters based on the SOM learning rule, where the winner node is determined by the maximum likelihood method. The second is a step in which the map structure adapts to the given data by determining which models are no longer neighbors, and which models are unnecessary using the MDL criterion.

\subsection{SOM based on the ML method}
Let $\bm{x}_i=(x_{i1},\,x_{i2},\,\ldots ,\,x_{ip})^t$ be $p$-dimensional data, and $\bm{X} = (\bm{x}_1\,\bm{x}_2\,\cdots\,\bm{x}_n)^t$ be a dataset containing $n$ samples. Let $\bm{M} = \{1,\,2,\,\ldots,\,M\}$ be a set of $M$ probability distribution models, and let $\bmt _m~(m\in \bm{M})$ be the model parameter of the $m$th model.
Assume that each sample follows one of the $M$ models independently. Let $m_i\in \bm{M}$ be the model number to which $\bm{x}_i$ belongs, and let $\bmT = \{\bmt _1,\,\bmt _2,\,\ldots ,\,\bmt _M\}$ be the collection of model parameters. Then, the likelihood considered in this study is described by
\begin{equation}
 L_C(\bmT) = \prod _{i=1}^n f(\bm{x}_i\mid \bmt _{m_i}).
\end{equation}
This is sometimes known as the {\it classification likelihood} in a classification context, or as the {\it complete-data likelihood} within the EM framework~\cite{fraley2002model}\cite{mclachlan2016mixture}. In this study, we estimate not only $\bmT$, but also $m_i$, which is the classification of sample $i$. Note that an estimated value of $m_i$, represented by $\hat{m}_i$, is a discrete value, namely, let $\hat{m}_i\in \bm{M}$.

Next, we describe an extension of Kohonen's SOM that assigns input samples to clusters using the maximum likelihood method. In this extension, a node represents one of the models $\bm{M}$, in which each sample of the dataset $\bm{X}$ belongs to only one of these models $\bm{M}$. Therefore, $\bm{x}_i$ is given, and the likelihood of each model of $\bm{M}$ can be calculated. Hence, the winner node is determined as the node with the maximum likelihood for a given sample $\bm{x}_i$ among the nodes as follows:
\begin{equation}
 \hat{m}_i = \argmax _{m\in \bm{M}} f(\bm{x}_i\mid \bmt _m),\label{223711_25Jul17}
\end{equation}
after which the winner node and its neighbor are adapted for $\bm{x}_i$ based on the SOM learning rule.

The adaptation is performed as follows. In this version of the SOM, we approximate the $k$th-order moments of
$\bm{x}$,
\begin{equation}
 E\left (\prod_{j=1}^p x_j^{r_j}\right )\quad {\rm where}\ \sum _{j=1}^p
  r_j = k, r_j = 0,1,\ldots ,k\label{002120_11Jul17}
\end{equation}
using a stochastic approximation method~\cite{robbins1951stochastic}. Let $\mu _{r_1\,r_2\,\cdots \,r_p}$ be a $k$th sample moment, where $r_j$ is a non-negative integer that satisfies $\sum _{j=1}^p r_j =k$. Under the mean squared error criterion between $\mu _{r_1\,r_2\,\cdots \,r_p}$ and (\ref{002120_11Jul17}), the update rule is given by
\begin{equation}
 \Delta_\tau \mu _{r_1\,r_2\,\cdots \,r_p} = \alpha (\tau) \left(\prod _{j=1}^p x_{j}^{r_j} -  \mu _{r_1\,r_2\,\cdots\,r_p}\right)\\\label{181819_16Jul17}
\end{equation}
where $\alpha (\tau)$ is the learning rate at time step $\tau$, and $0 < \alpha (\tau) < 1$ and decreases monotonically.

This moment approximation rule provides a simple parameter update rule for some probability distributions, where the parameters can be estimated using the method of moments. 

We call this extension of the SOM the maximum likelihood SOM (MLSOM) to distinguish it from Kohonen's SOM. We present MLSOM for continuous and count data using the Gaussian and the multinomial model, respectively, in this paper.

\subsubsection{Gaussian model}
Consider, for instance, the $p$-dimensional normal distribution
\begin{eqnarray}
 \lefteqn{f(\bm{x}\mid \bm{\mu},\,\bm{\Sigma})=} \nonumber \\
  &&
  \frac{1}{(2\pi)^{p/2}|\bm{\Sigma}|^{1/2}}\exp
  \left\{-\frac{1}{2}(\bm{x}-\bm{\mu})^t\bm{\Sigma} ^{-1}(\bm{x}-\bm{\mu})\right\}
\end{eqnarray}
where $\bm{\mu}$ is a mean vector, and $\bm{\Sigma}$ is the covariance matrix. Based on the approximation rule (\ref{181819_16Jul17}) and using the method of moments, a sample $\bm{x}_i$ assigned the parameters of the $m$th node is updated as follows:
\begin{eqnarray}
 \Delta_\tau \bm{\mu}_m & = & \alpha(\tau) (\bm{x}_i - \bm{\mu}_m)\label{171320_24Jan17}\\
 \Delta_\tau \bm{\Sigma}_m & = & \alpha(\tau)[(1-\alpha(\tau))(\bm{x}_i-\bm{\mu}_m)(\bm{x}_i-\bm{\mu}_m)^t - \bm{\Sigma}_m]\label{151915_24Sep21}
\end{eqnarray}
where $\bm{\mu}_m$ is a $p$-dimensional real vector, and $\bm{\Sigma}_m$ is a real symmetric matrix of size $p\times p$. The derivation is presented in \ref{gaussian model}.

The MLSOM is a generalization of Kohonen's SOM. Consider a $p$-dimensional normal distribution in which the covariance matrix is the identity matrix, then the log-likelihood is proportional to $-\frac{1}{2}(\bm{x}-\bm{\mu})^t(\bm{x}-\bm{\mu})$. Hence, the rule for finding the winner node (\ref{223711_25Jul17}) is to minimize the Euclidean distance between $\bm{x}_i$ and $\bm{\mu}_m$. In this case, it is no longer necessary to update $\bm{\Sigma}$, and only the parameters must be updated (\ref{171320_24Jan17}). Therefore, the MLSOM coincides with Kohonen's SOM, as described in Section \ref{sec:background}.

\subsubsection{Multinomial model}
\label{multinomial model}
Suppose $\bm{x} = (x_1,\,x_2,\,\ldots ,\,x_p)^t$ follows a
multinomial distribution. The probability function is given by
\begin{equation}
 f(\bm{x}\mid \bmt) = \frac{(\sum _{j=1}^p x_j)!}{x_1!\,x_2!\,\cdots
  x_p!}\prod _{j=1}^p \theta _j^{x_j},
\end{equation}
where $\theta _j \geq 0,\,\sum _{j=1}^p\theta _j = 1$. The 1st moment of
the multinomial distribution is given by
\begin{equation}
 E(\bm{x}) = \left(\sum _{j=1}^p x_j\right)\bmt.
\end{equation}

The procedure for updating the parameters of the multinomial model in MLSOM, based on Eq.~(\ref{181819_16Jul17}), is expressed as follows:
\begin{equation}
 \Delta _\tau \bmt _m = \alpha(\tau)\left(\frac{\bm{x}_i}{\sum _{j=1}^p
             x_{ij}} - \bmt _m\right).
\end{equation}
Note that if $\sum _{j=1}^p x_{ij}=0$, then let $\Delta _\tau \bmt _m= \bm{0}$.

\subsection{Map structure update}

The method described in this section to update the map structure is comprised of two components: disconnecting weak links, and deleting unnecessary nodes. We introduce some notation to explain this method. Let $\bm{B} \subseteq \{\{i,\,j\}\mid i,\,j\in \bm{M}\}$ be a set of undirected edges that are two-element subsets of a set of nodes $\bm{M}$, and a map is represented by a graph $(\bm{M},\,\bm{B})$. The elements of $\bm{B}$ represent links that represent the neighborhood relationships between nodes.

\subsubsection{Link cutting}
Consider the problem of determining whether an edge $\{m,\,l\}\in \bm{B}$ is removed. Here, we measure the weakness of a node connection by using the KL divergence~\cite{kullback1951information}. Let $D_{\rm KL}(f_{\bmt}\| f_{\bmt'})$ be the KL divergence for the two probability distribution models $f_{\bmt}=f(\cdot \mid \bmt)$, $f_{\bmt'}=f(\cdot \mid \bmt')$, defined as follows:
\begin{eqnarray}
  D_{\rm KL}(f_{\bmt}\|f_{\bmt'}) & = & E_{f_{\bmt}}\left[\log \frac{f(\bm{x}\mid \bmt)}{f(\bm{x}\mid \bmt')}\right] \nonumber \\
 & = & \int _{\bm{x}} f(\bm{x} \mid \bmt) \log \frac{f(\bm{x}\mid \bmt)}{f(\bm{x}\mid \bmt')} d\bm{x},
  \label{eq:kl-div}
\end{eqnarray}

Let $\hat{D}(m,\,l)$ be the weakness of the connection $\{m,\,l\}$ as defined by
\begin{equation}
 \hat{D}(m,\,l) = \frac{1}{2} \hat{D}_{\rm KL}(f_{\bmt _m}\| f_{\bmt _l}) + \frac{1}{2} \hat{D}_{\rm KL}(f_{\bmt _l}\| f_{\bmt _m}),
\end{equation}
where $\hat{D}_{\rm KL}(f_{\bmt_m}\| f_{\bmt_l})$, which is an estimator of the KL divergence, is defined as follows:
\begin{equation}
 \hat{D}_{\rm KL}(f_{\bmt_m}\|f_{\bmt_l}) = \frac{1}{|\bm{S}_m|}\sum _{i \in \bm{S}_m}  \log \frac{f(\bm{x}_i\mid \bmt_m)}{f(\bm{x}_i\mid \bmt_l)},
  \label{eq:est-kldiv}
\end{equation}
where $\bmS _m = \{i\mid m_i=m,\,\forall\,i=1,\,2,\,\ldots,\,n\}$. Eq.~(\ref{eq:est-kldiv}) represents the quantity of likelihood deterioration for each
sample when all samples belonging to node $m$ move to node $l$; $\hat{D}_{\rm KL}(f_{\bmt_l}\| f_{\bmt_m})$ is defined similarly.

The threshold for $\hat{D}(m,\,l)$ is used to calculate the average likelihood for each node.

\begin{equation}
 \hat{D}_m=\frac{1}{|\bm{S}_m|} \sum _{i \in \bm{S}_m} \log f(\bm{x}_i\mid \bmt_m),
\end{equation}
The following rule is then used to determine whether to remove the edge $\{m,\,l\}$:
\begin{equation}
 \hat{D}(m,\,l) > \beta h,\quad {\rm where}\ h = \max _{m \in \bm{M}} (-\hat{D}_m),
  \label{eq:weak connect detection}
\end{equation}
where the parameter $\beta \geq 0$ controls the hardness to remove
edges. The threshold $h$, which represents the worst likelihood among the nodes, makes it difficult to cut the edges. If many isolated
nodes without edges to others exist, the SOM learning rule reduces to
the simple competitive learning rule, in which case the learning process may converge to a
poor local optimum. Therefore, it is preferable to retain edges as much as
possible to avoid a poor local optimum.

\subsubsection{Node deletion}
\label{node deletion}
The node deletion procedure determines whether to remove a node. An unnecessary node is determined based on the MDL criterion to remove it from the graph. In our setting, we cannot encode all the samples without specifying the model with which each sample is encoded~\cite{10.5555/559448}. Therefore, we need to encode samples and models on a one-to-one basis ${m_1,\,m_2,\,\ldots,\,m_n}$ to ensure that they correspond. Because the probability distribution of $m_i$ cannot be known in advance, assuming the probability of $m_i=m$ is $1/|\bm{M}|,\,m\in \bm{M}$, the codelength of $\{m_1,\,m_2,\,\ldots ,\,m_n\}$ is $n\log |\bm{M}|$. Note that it is also possible to include $\{m_i\}_{i=1}^n$ in the model itself and consider its degrees of freedom (see \cite{HOFMEYR2020106974} for K-means), but we do not treat them as such here for simplicity.

All samples are then classified into $|\bm{M}|$ groups using the information of $\{m_i\}_{i=1}^n$, and each sample group is encoded based on the corresponding model.

Therefore, we consider the following MDL form:
\begin{eqnarray}
 \lefteqn{{\rm MDL}(\bmM,\,\mathcal{S},\,\hat{\bmT})} \nonumber \\
 & = & \sum_{m\in \bm{M}} \left[-\sum_{i\in \bmS_m} \log f(\bm{x}_i\mid \hat{\bmt}_m) \right] + \frac{{\rm df}(\hat{\bmT})}{2}\log n +
   n\log |\bmM| \nonumber \\
  & = & -\log L_C(\hat{\bmT}) + \frac{{\rm df}(\hat{\bmT})}{2}\log n +
   n\log |\bmM|,
   \label{cmdl}
\end{eqnarray}
where $\mathcal{S}=\{\bmS_m\}_{m\in\bm{M}}$, which specifies the partition of samples $\bmS _m = \{i\mid m_i=m,\,\forall\,i=1,\,2,\,\ldots,\,n\}$. Minimizing Eq.~(\ref{cmdl}) over every $\mathcal{S}$ enables the optimal classification $\hat{\mathcal{S}}$ to be obtained. Consequently, the MDL of Eq.~(\ref{cmdl}) also includes an assessment for the classification of samples, namely, the clustering result, unlike the standard MDL.

Using this MDL evaluation, for each $m\in \bm{M}$, an unnecessary node is determined as follows:
\begin{enumerate}
 \item Evaluate the current map $(\bm{M},\bm{B})$ using the MDL (\ref{cmdl}).
 \item Classify each sample of $\bm{S}_m$ to a node of $\bm{M}-\{m\}$ by the maximum likelihood method.
       \label{184636_12Aug17}
 \item Using the above-classified samples, estimate the parameters of nodes $\bm{M}-\{m\}$ by the method of moments.
 \item Let the map of nodes $\bm{M}-\{m\}$ with new parameters be a candidate map, and evaluate the map using the MDL.
       \label{184644_12Aug17}
 \item Execute \ref{184636_12Aug17}--\ref{184644_12Aug17} for all nodes, and select the map with the best MDL among the candidates.
 \item Compare the MDL of the selected candidate map with the current map, and select the one that is more appropriate.
\end{enumerate}

If the node is deleted, the edges are removed to the deleted node. In addition, for each node that was adjacent to the deleted node, new edges are inserted between all these nodes to prevent isolated nodes from being generated for the abovementioned reason.

\subsection{Complete algorithm}
\begin{algorithm*}[t]
 \centering
 \caption{SMLSOM (dataset $\bm{X}$, edge hardness $\beta$, map $(\bm{M},\bm{B})$)}
 \label{SMLSOM algorithm}
 \begin{algorithmic}[1]
  \State Let $(\bm{M}^{(0)}, \bm{B}^{(0)}) \leftarrow (\bm{M},\bm{B})$. Initialize the node with parameter $\bmt _m\,(m\in \bm{M}^{(0)}$).
  \State Let $t \leftarrow 0$
  \Repeat
  \State Run MLSOM on $(\bm{M}^{(t)}, \bm{B}^{(t)})$.
  \State Identify weak connections using (\ref{eq:weak connect detection}) with edge hardness $\beta$ and remove those from $\bm{B}^{(t)}$ if they exist.
  \State Determine which nodes are unnecessary based on the MDL (\ref{cmdl}) and remove it from $\bm{M}^{(t)}$ if it exists. Upon deletion of a node, the neighbors of the deleted node are restored, as described in Section~\ref{node deletion}.
  \State Let $(\bm{M}^{(t+1)}, \bm{B}^{(t+1)}) \leftarrow (\bm{M}^{(t)}, \bm{B}^{(t)})$.
  \State Let $t \leftarrow t + 1$.
  \Until {map structure not changed}
  \State Return $(\bm{M}^{(t)}, \bm{B}^{(t)})$.
 \end{algorithmic}
\end{algorithm*}

The SMLSOM algorithm is shown in Algorithm~\ref{SMLSOM
algorithm}. The initial map structure was selected as a rectangular or hexagonal lattice graph. The map can be initialized in two ways. The first approach is to initialize all the parameters randomly. Second, only the first moments, namely, the mean vectors, are initialized based on the principal component of dataset $\bm{X}$ (see \ref{Init PCA}),
and higher moments are initialized uniformly or randomly. We recommend the second approach because
this form of initialization is likely to produce similar maps, even if the order of inputs is different.

Note that we chose a simple function of Eq.~(\ref{eq:neighborhood function_2}) as the neighborhood function $h_{c\,m}$ for simplicity. In SMLSOM, the map is represented as a graph of nodes and links, so the distance between nodes is not the Euclidean distance. $d(c, m)$ is the length of the shortest path between node $c$ and node $m$, and the neighborhood radius $r(\tau)$ is the threshold for how distant nodes are considered a neighbor\footnote{The Kohonen package, which is a SOM library for R, uses the scheduling scheme $r(\tau)=r_1-(r_1-r_2)\,\tau/\tau_{\rm max}$. Note that $r(\tau)=0.5$ is set when $r(\tau)<1$ (i.e., only itself is updated). The $\tau_{\rm max}$ is the total number of iterations of the algorithm to be set in advance, $r_1>0$ is the initial value of the neighborhood radius, and $r_2=-r_1$. Depending on the value of $r_1$, $r(\tau)=0.5$ (no neighborhood) is obtained at about 1/3 of $\tau_{\rm max}$.}. The $\alpha(\tau)$ is usually set to linearly decay from 0.05 to 0.01.

In the MLSOM, clusters are constructed using the maximum likelihood method with {\it
soft-to-hard} learning, which is the SOM learning rule that uses the neighborhood relationship between nodes defined by a graph, $(\bm{M},\,\bm{B})$. The link-cutting procedure removes the edges between dissimilar nodes. Consequently, in the updated version of the SOM, the nodes connected by the remaining links are updated efficiently to similar ones. In the node deletion procedure, when two similar nodes exist, one of the nodes is deleted such that it is merged with the other. New edges are then added between nodes that are adjacent to the deleted node. This procedure prevents the neighborhood relationship between nodes from excessively changing when a node is removed from a map. Thus, when the mended map is input into the MLSOM, the nodes that are adjacent to the deleted node are organized. Hence, the learning process of the SMLSOM to reduce the number of clusters can proceed.

\subsection{Computational complexity}
The SMLSOM algorithm has three components: MLSOM, link cutting, and node deletion. MLSOM requires $M$ comparisons to make a sample belong to one of the $M$ clusters. For all the samples, $nM$ calculations are required. In link cutting, at most $n$ additions are required to calculate the weight of one link. If we consider an undirected graph with $M$ nodes, the number of links is at most $M(M-1)/2$, so that the amount of computation required is at most $nM(M-1)/2$. In node deletion, assigning the samples belonging to the target node to other nodes requires at most $n(M-1)$ computations. After executing this reassigning for each $M$ node, the required computation is at most $nM(M-1)$. Therefore, the amount of calculation for a given $M$ is at most $3nM(M-1)/2+nM$. If this calculation is performed while decreasing $M$ one-by-one with $\beta=\infty$, the overall result will be $O(nM^3)$ from the formula of the sum of series.

\section{Experiment}
\label{sec:experiment}

In this section, we clarify the effectiveness of the proposed method compared with other methods. The remainder of this paper is organized as follows. 

First, we demonstrate that the learning process of SMLSOM can be used to determine a suitable number of clusters, and we compare its characteristics with those of other methods using small real datasets. 

Second, we evaluate the selection of $M$, the accuracy of the clustering, and the computational time with SMLSOM in comparison with the other selected methods. We use X-means~\cite{pelleg2000x}, G-means~\cite{hamerly2004learning}, PG-means~\cite{feng2007pg}, Dip-means~\cite{kalogeratos2012dip},
MML-EM~\cite{figueiredo2002unsupervised}, and Mclust~\cite{fraley2007model}\cite{RJ-2016-021}. The artificial dataset used in the simulation is generated using the MixSim package\footnote{MixSim: Simulating Data to Study Performance of Clustering Algorithms. Available at \url{https://cran.r-project.org/web/packages/MixSim/index.html}}~\cite{melnykov2012mixsim} in R, which can vary the overlap between clusters. The experimental results show that SMLSOM delivers high performance within a short computational time.

Finally, to demonstrate the applicability of the proposed method, we apply SMLSOM to continuous and count data. In the former, we deal with applying high-dimensional data using an image dataset. The latter type of count data is more advantageous for the introduction of a probabilistic model. In actual count data, phenomena such as overdispersion (see \cite{CORSINI2022107447} for details) are observed, and appropriate handling is required. For example, a probability model, such as a negative binomial distribution or a zero-inflated Poisson model, can deal with these phenomena.

It is not helpful for applications if a method selects $M$ correctly, but an uninterpretable result is obtained. Therefore, we also analyze the clusters produced with SMLSOM and discuss whether they can be interpreted and contain the appropriate data.

\subsection{Demonstration of the algorithm using real data}

\begin{figure*}[!t]
 \centering
 \includegraphics[bb=50 0 1080 432,scale=.4]{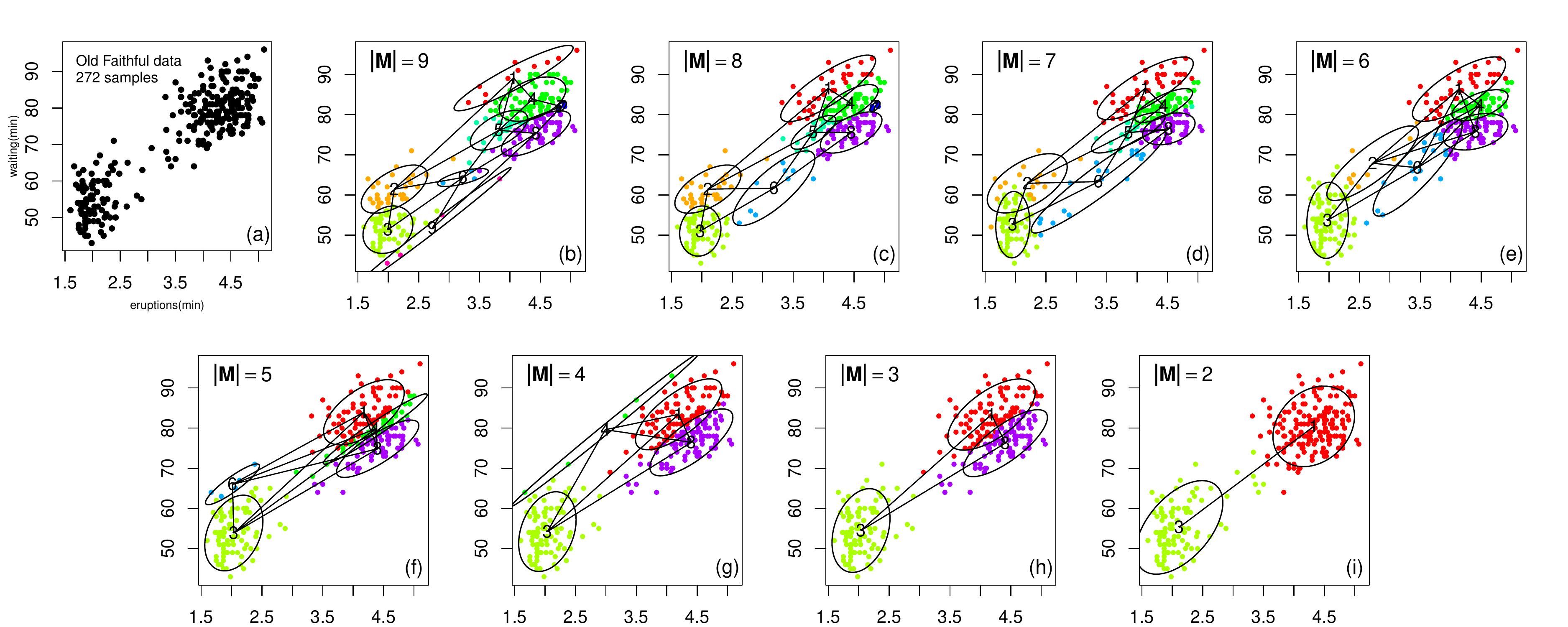}
 \caption{Fitting the Old Faithful dataset using SMLSOM: (a) original data, (b)--(i) estimates obtained after each iterative cycle of the algorithm. The algorithm
 starts with $|\bm{M}|=9$ and terminates at $|\bm{M}|=2$. Each sample is colored corresponding to the cluster to which it belongs. The solid ellipses represent the level curve of each node estimate. The solid lines represent the links between the nodes. The numbers in the center of an ellipse are the numbers
 of nodes that have not been deleted in the particular iterative cycle.}
 \label{002744_7May17}
\end{figure*}

In this section, we present a demonstration of the proposed method using a real dataset and compare it with existing methods regarding the selection of a model for clustering.

We consider the Old Faithful dataset to fit bivariate Gaussians with full covariance matrices. For comparison, we run SMLSOM and Mclust using this dataset as input. We then compare the results obtained with those reported previously by other studies. For SMLSOM, we start with a $3\times 3$ hexagonal lattice (i.e., $M=9$) using principal component analysis (PCA) initialization, the initial covariance matrix of each node was set to the identity matrix, and $\beta = 15$. For Mclust, we use $M=1$ to $M=9$ and evaluate the results using BIC/ICL. The default initialization method of the EM algorithm in Mclust is based on model-based hierarchical agglomerative clustering (see \cite{fraley2007model}, \cite{fraley2002model}). Fig.~\ref{002744_7May17} shows the intermediate estimates and the final estimate ($\hat{M}^*=2$) produced by SMLSOM. The figure shows the ``shrinking'' process of the map as the map structure is updated by deleting a node and cutting links. With SMLSOM, $\hat{M}^*=2$ is selected 99 times with 100 runs; hence, the two clusters are strongly supported. With Mclust, similar to the results of SMLSOM, two components are selected with both BIC and ICL. The top three BIC results are $-2322.2 (M=2)$, $-2333.9 (M=3)$, and $-2359.2 (M=4)$. Similarly, the ICL results are $-2322.7 (M=2)$, $-2361.9 (M=3)$, and $-2464.4 (M=4)$. Therefore, the results of Mclust support two clusters for the dataset. On the other hand, the results obtained with the Bayesian inference method~\cite{richardson1997bayesian} contrast with the abovementioned results, which support two clusters. That is, \cite{dellaportas2006multivariate} reported that the posterior probability of the three components is the highest (0.5845) and two components are the 2nd highest (0.3035), while another study \cite{komarek2009new} found that three and two components have almost the same posterior probabilities.

Thus, the appropriate number of clusters varies depending on the algorithm used, even with the same data. The ``true'' number of clusters does not exist in actual data, and the choice of a reasonable number of clusters depends on the task. Therefore, the consistency of selection is more important for the algorithm than for assuming and estimating the ``true'' number of clusters.

\subsection{Simulation with artificial data}
In this artificial data experiment, we will validate which algorithm performs better for various ``difficulty level'' cases in estimating the number of clusters. 

We use the Gaussian mixture model as the data generation model. The parameters of the experiment are the following six parameters; data size $n$, number of dimensions $p$, the true number of clusters $M^*$, the shape of the covariance matrix (e.g., diagonal or full covariance), mixing probabilities $\pi_m$, and degree of overlap between the clusters.

The difficulty of estimating the number of clusters is mainly determined by the cluster overlap. In this case, the number of dimensions $p$ does not determine the difficulty of estimating the number of clusters. It is not a hard task to separate data if the data are linearly separable, even in high dimensionality. Also, changing $M^*$ is not essential. Since a Gaussian mixture model generates the samples, all clusters are Gaussians; there is no need to identify them by name. Unless the data generation model differs fundamentally from cluster to cluster (e.g., cluster A is a Gaussian distribution and cluster B is a multinomial distribution), the distribution of the estimated cluster number is determined by the cluster overlap under a given $M^*$ rather than by $M^*$. This fact can be imagined from, for example, the case that if the clusters overlap entirely, the best estimate will be smaller than $M^*$ since there is no way to distinguish between them.

Although we think it also would be interesting to investigate the scenarios when varying the data size $n$, the shape of the covariance matrix, and the mixing probabilities $\pi_m$, the number of trials would be too large, so we use a simple setup in here.

\subsubsection{Cluster overlap}
We generate an artificial dataset using MixSim, which considers the following:
\begin{equation}
  \begin{array}{r}
   \omega _{l\mid m} ={\rm Pr}[\pi _m f(\bm{x}\mid
   \bm{\mu}_m,\,\bm{\Sigma}_m) < \pi _l f(\bm{x} \mid \bm{\mu}_l,
   \,\bm{\Sigma}_l)], \\
   {\rm where}\ \bm{x}\sim \mathcal{N}(\bm{\mu}_m,\,\bm{\Sigma}_m),
  \end{array}
\end{equation}
where $\omega _{l\mid m}$ is the misclassification probability that sample $\bm{x}$ generated from the $m$th component was classified mistakenly to the $l$th component, and $\omega _{m\mid l}$ is defined similarly. The overlap between two components is defined by
\begin{equation}
  \omega _{m\,l} = \omega _{m\mid l} + \omega _{l\mid m}.
\end{equation}
We can specify $\bar{\omega}$, the average of $\omega_{m\,l}$, and generate datasets using MixSim. The procedure MixSim uses to generate the data corresponding with the overlap $\bar{\omega}$ is as follows~\cite{melnykov2012mixsim}:
\begin{enumerate}
 \item Mean vector $\bm{\mu}_m$ is sampled from a $p$-dimensional
       uniform distribution and covariance matrix $\bm{\Sigma}_m$ is
       obtained from the standard Wishart distribution with parameter $p$ and
       $p+1$ degrees of freedom. Note that, the user can specify
       the structure of covariance matrices as being either spherical or
       non-spherical, and heterogeneous or homogeneous. If the spherical
       structure is specified, $\bm{\Sigma}_m=\sigma_m \bm{I}$, and $\sigma_m$ is taken from the standard uniform distribution. If
       the homogeneous structure is specified, set $\bm{\Sigma}_1 =
       \bm{\Sigma} _2 = \cdots = \bm{\Sigma}_M = \bm{\Sigma}$, where
       $\bm{\Sigma}$ is generated by either of the aforementioned two methods (spherical or non-spherical).
 \item The covariance matrices are multiplied by a positive constant $c$,
       after which the value of $c$ that minimizes the difference between the user-specified $\bar{\omega}$ and $\hat{\bar{\omega}}$ is determined by
       the current multiplier $c$.
\end{enumerate}

\subsubsection{Evaluation}
\label{clause: evaluation}

For the artificial data in this section and image data in the next section, the adjusted Rand index~(ARI)~\cite{Hubert1985} and the normalized mutual information (NMI)~\cite{strehl2002cluster} were used as evaluation indices. For the artificial and image data, the samples are pre-labeled as to which class they belong to.

In the evaluation with ARI and NMI, we evaluated the pair of a given
label and the clustering obtained by the algorithm. ARI looks at the rate of agreement between both classifications by label and clustering. A value is high when a pair of samples belong to the same cluster if they belong to the same label and to different clusters if they belong to different labels. Thus, it is a lower value in cases where the number of groups in both labels and clustering does not match. On the other hand, NMI looks at the amount of mutual information between labels and clusterings. This metric evaluates the extent to which knowing the clustering result reduces uncertainty in given labels. Therefore, even if the number of groups in both the labels and clustering does not match, the NMI will take a high value when each cluster collects samples of one specific label. See \ref{ARI} and \ref{NMI} for detailed calculations.

In both experiments, we evaluated the results using the two indices. However, we adopted the index for the final decision, which is consistent with the experiment's objective.

\subsubsection{Comparison methods and its parameters settings}
\label{clause: parameter_setting}

\begin{table}[t]
    \centering
    \caption{Parameters of each algorithm. }
    \begin{tabular}{ll} \hline
        Algorithm & Parameters \\ \hline
        X-means & $M_{\rm max}=100$, $M_{\rm min}=1$ \\
        G-means & $M_{\rm max}=100$, $M_{\rm min}=1$, $\alpha = 10^{-4}$\\
        Dip-means & $M_{\rm max}=100$, $M_{\rm min}=1$, $\alpha = 10^{-16}$, $v_{thd}=0.01$\\
        PG-means & $M_{\rm max}=100$, $M_{\rm min}=1$, $\alpha = 10^{-3}$, $N_{\rm prj} = 12$, $\epsilon = 10^{-4}$\\
        MML-EM & $M_{\rm max}$, $M_{\rm min}=1$, $\epsilon = 10^{-4}$\\
        Mclust & $M_{\rm max}$, $M_{\rm min}=1$\\
        SMLSOM & $M_{\rm max}$, $M_{\rm min}=1$, $\beta = 15$, $\alpha(\tau)$, $r(\tau)$, $\tau_{\rm max}$\\ \hline
        & \multicolumn{1}{r}{\footnotesize $M_{\rm max}$: maximum number of estimated clusters}\\
        & \multicolumn{1}{r}{\footnotesize $M_{\rm min}$: minimum number of estimated clusters}\\
        & \multicolumn{1}{r}{\footnotesize $\alpha$: significance level} \\
        & \multicolumn{1}{r}{\footnotesize $v_{thd}$: split viewer ratio (see text)} \\
        & \multicolumn{1}{r}{\footnotesize $N_{\rm prj}$: number of projections}\\
        & \multicolumn{1}{r}{\footnotesize $\epsilon$: relative convergence tolerance for the EM algorithm}\\
    \end{tabular}
    \label{tab:params_methods}
\end{table}

We used existing methods for the experiment. Mclust\footnote{mclust: Gaussian Mixture Modelling for Model-Based Clustering, Classification, and Density Estimation \url{https://cran.r-project.org/web/packages/mclust/index.html}} is a Fortran implementation of the EM algorithm for Gaussian mixtures. The MATLAB code of MML-EM\footnote{The MATLAB code is available at \url{http://www.lx.it.pt/~mtf/}, accessed 2022/9} is published by the author of this method was referred to in our experiment. In general, MATLAB is inferior to C and Fortran in terms of the computational time to complete a loop procedure; thus, we converted the MATLAB code into C to enable us to compare the computational time with other Fortran and C implementations of the methods. We implemented other methods, X-means~\cite{pelleg2000x}, G-means~\cite{hamerly2004learning}, PG-means~\cite{feng2007pg}, and Dip-means~\cite{kalogeratos2012dip} in C based on those papers. We published codes of the methods at: \url{https://github.com/lipryou/searchClustK}

The parameters of each algorithm are shown in Table~\ref{tab:params_methods}. Note that $M_{\rm max}$ and $M_{\rm min}$ are the range of the cluster number search. In X-means, G-means, PG-means, and Dip-means, its search start from $M_{\rm min}$. Unless otherwise noted, $M_{\rm min}=1$ and $M_{\rm max}=100$ were set in these methods. Because $M_{\rm max}$ of those methods does not affect the computation time if $M_{\rm max}$ is sufficiently large. On the other hand, in MML-EM, Mclust, and SMLSOM, $M_{\rm max}$ should be set according to the situation. $\alpha$ is the significance level of each statistical test. $v_{thd}$ is a threshold related to cluster splitting. Dip-means executes the unimodality test for each sample in a cluster and calculates the proportion of significant samples within the cluster (called split viewer ratio). The cluster will be divided if the split viewer ratio is larger than $v_{thd}$. $N_{\rm prj}$ is the number of projections in PG-means. We adopted the values of $\alpha$, $v_{thd}$ and $N_{\rm prj}$ recommended in proposed papers. $\epsilon$ is a threshold of the EM algorithm convergence determination. $\epsilon=10^{-4}$ was adopted from the MML-EM MATLAB code. In SMLSOM, the learning rate $\alpha(\tau)$ and the neighborhood radius $r(\tau)$ were set according to the defaults in the Kohonen package\footnote{Kohonen: supervised and unsupervised SOMs. Available at: \url{https://cran.r-project.org/web/packages/kohonen/index.html}} in R. $\tau_{\rm max}$ was set the data size $n$ unless otherwise specified. $\beta$ is discussed at the end of this section.

\subsubsection{Result}
\begin{figure}[t]
 \centering
 \hspace*{-1cm}
 \begin{tabular}{cc}
  \includegraphics[bb=0 0 432 360, width=7cm]{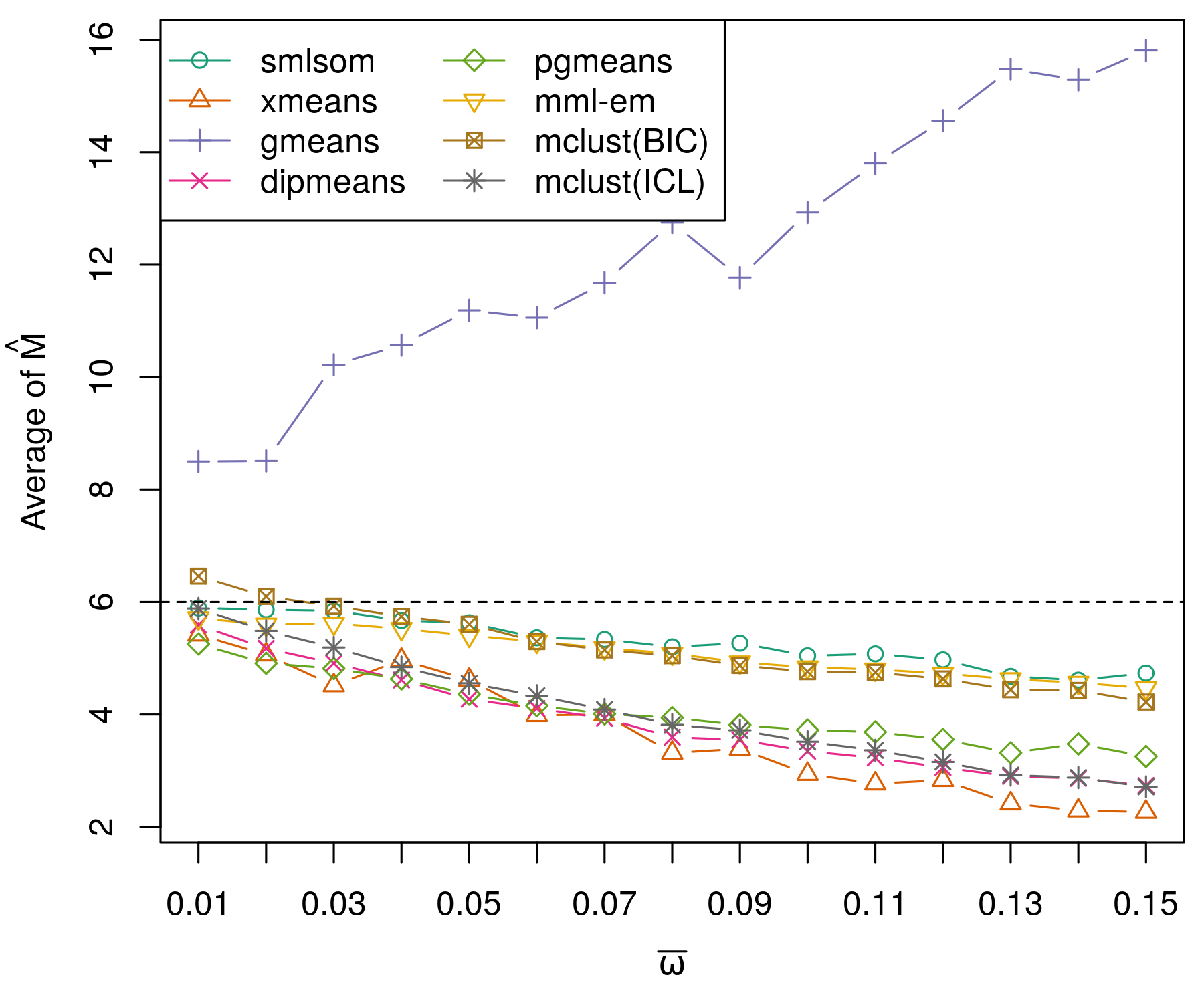} &
  \includegraphics[bb=0 0 432 360, width=7cm]{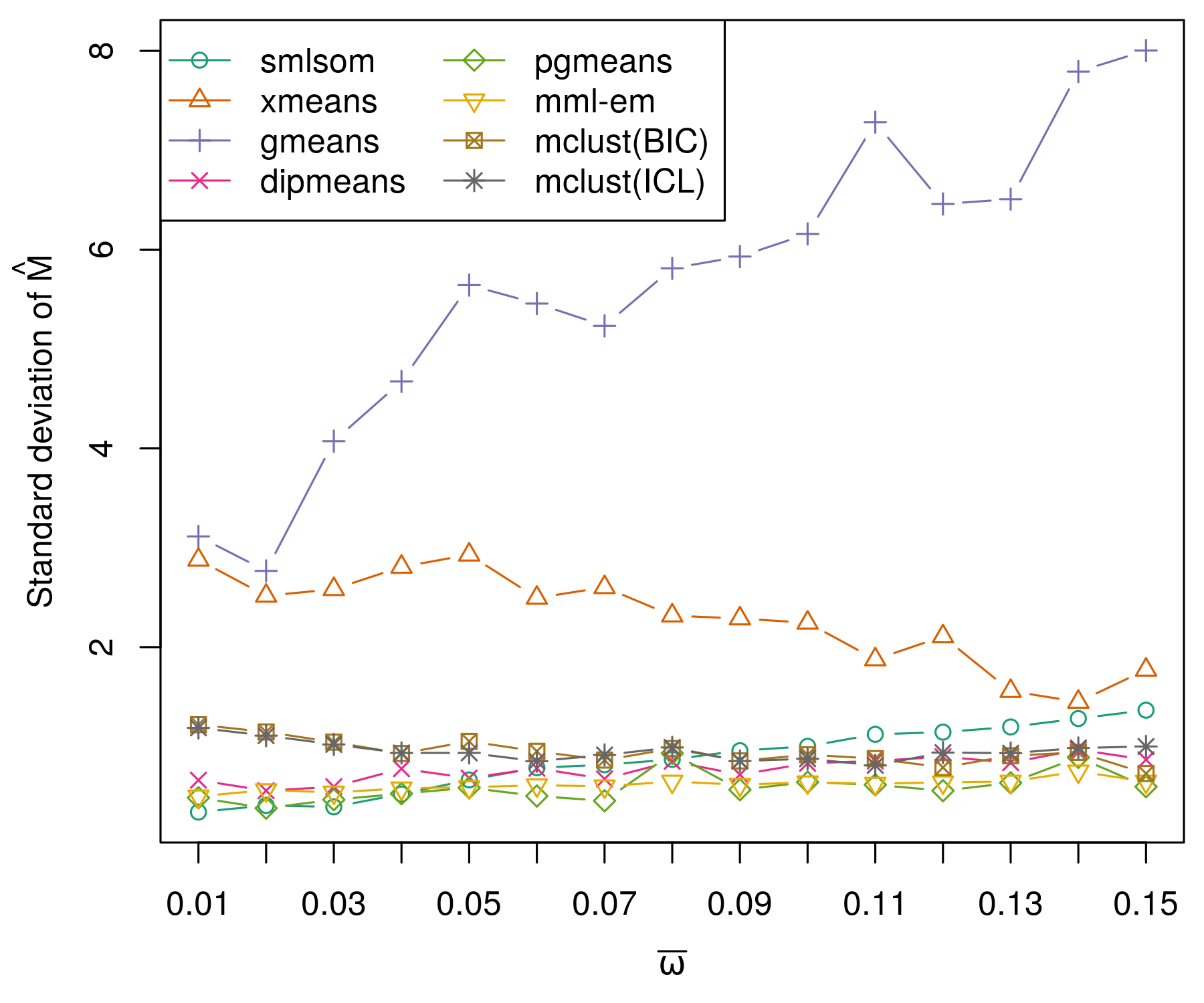} \\
  (a) Average &
  (b) SD
 \end{tabular}
 \caption{Evaluation for the number of clusters selected over 1000 simulations: (a) Averages (black dash line indicates reference value $M^*$), (b) Standard deviations.}
 \label{fig:selections behavior}
\end{figure}

We generated samples from the GMM with spherical and heterogeneous covariance matrices using MixSim. The sample was $p=2$, and the number of samples was $n=3,000$. The number of components was $M^*=6$, and the mixing probability of each component was $\pi_m = 1/M^*$. We set $\bar{\omega}$ to 15 different values and created 100 sets of samples with the abovementioned conditions for each $\bar{\omega}$. Each method was run 10 times for each sample set. Thus, 1000 values of $\hat{M}^*$ were produced by each method for each $\bar{\omega}$.

The settings of each algorithm are as follows. The initial parameters were randomly initialized. The covariance matrices estimated by PG-means, MML-EM, Mclust, and SMLSOM were of the full covariance type. MML-EM started with $M_{\rm max}=9$ and SMLSOM also started with a $3\times 3$ hexagonal lattice~(i.e., $M_{\rm max}=9$). Mclust was applied to each of $M_{\rm min}=1$ to $M_{\rm max}=9$, and selected the best result evaluated by the BIC and ICL.

We evaluated the results of the methods from three points of view: the behavior of the estimation $\hat{M}$, the accuracy and stability of clustering, and the computational time. Clustering accuracy was measured by the average of ARI or NMI. Stability was evaluated by determining the standard deviation of ARIs (NMIs). Note that the computational time of Mclust was measured as the total execution time required to determine $\hat{M}^*$.

We ran all methods on a computer running Ubuntu 22.04, with two Xeon SC 4208 8C 2.1GHz, and 128-GB memory. Fig.~\ref{fig:selections behavior} shows the bias and variance of the selection by each method. Fig.~\ref{fig:selections behavior}(a) shows that when the reference value is the number of distributions ($M^*=6$), SMLSOM can estimate with the least bias, on average. On the other hand, Fig.~\ref{fig:selections behavior}(b) shows that the SMLSOM estimate $M^*$ with the lowest variance when $\bar{\omega}$ is low, but the variance increases as $\bar{\omega}$ increases. When $\bar{\omega}$ is high, the distributions overlap significantly and may not be distinguishable as clusters. In this case, the algorithm also considers $M=5$ and $M=4$ as candidates for selection. As shown in Fig.~\ref{fig:selections behavior}(a), estimated values move away from the reference value to the lower side as $\bar{\omega}$ increases for each method, except G-means.

\begin{figure*}[t]
 \centering
 \hspace*{-1cm}
 \begin{tabular}{cc}
  \includegraphics[bb=0 0 432 360, width=7cm]{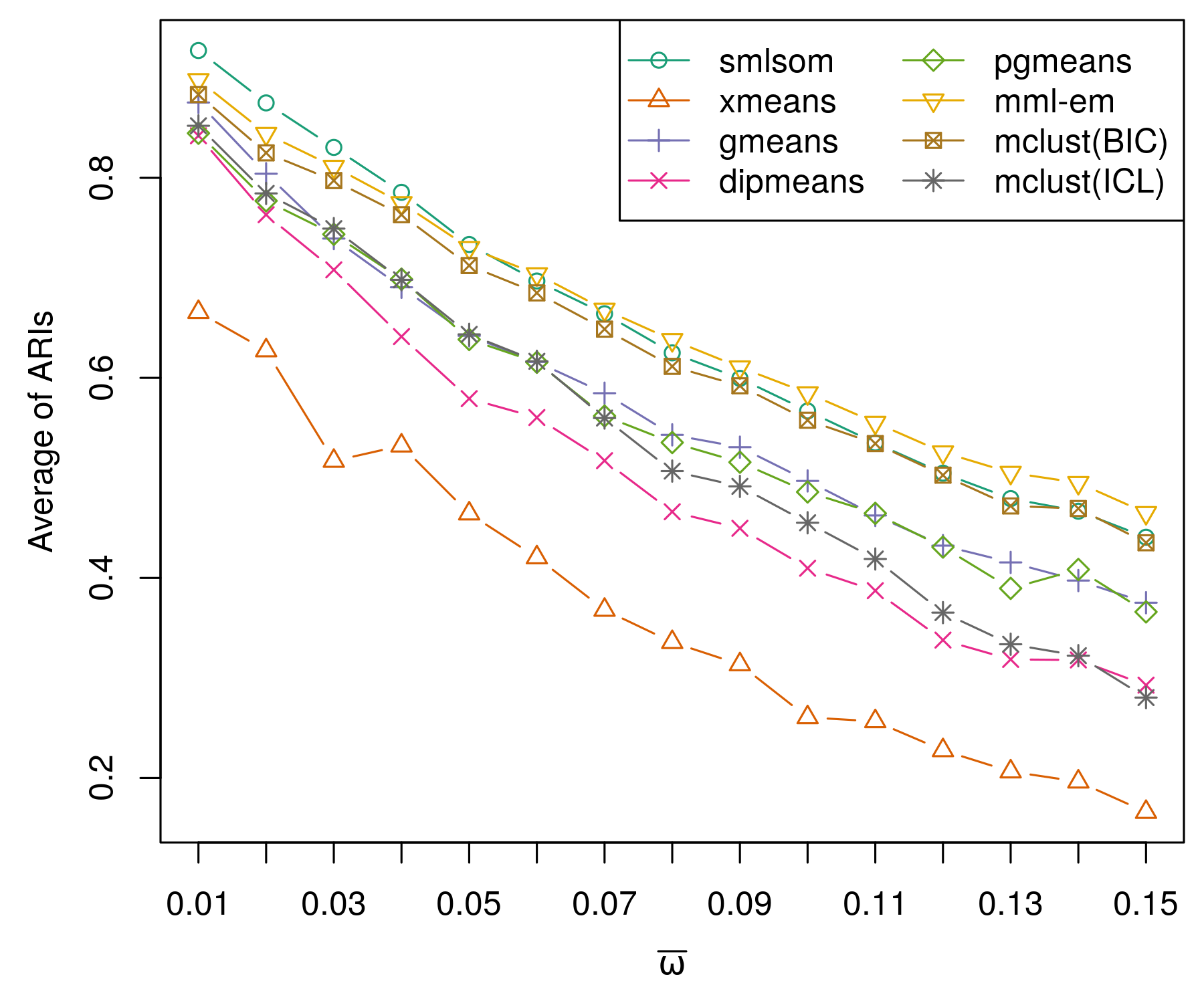} &
  \includegraphics[bb=0 0 432 360, width=7cm]{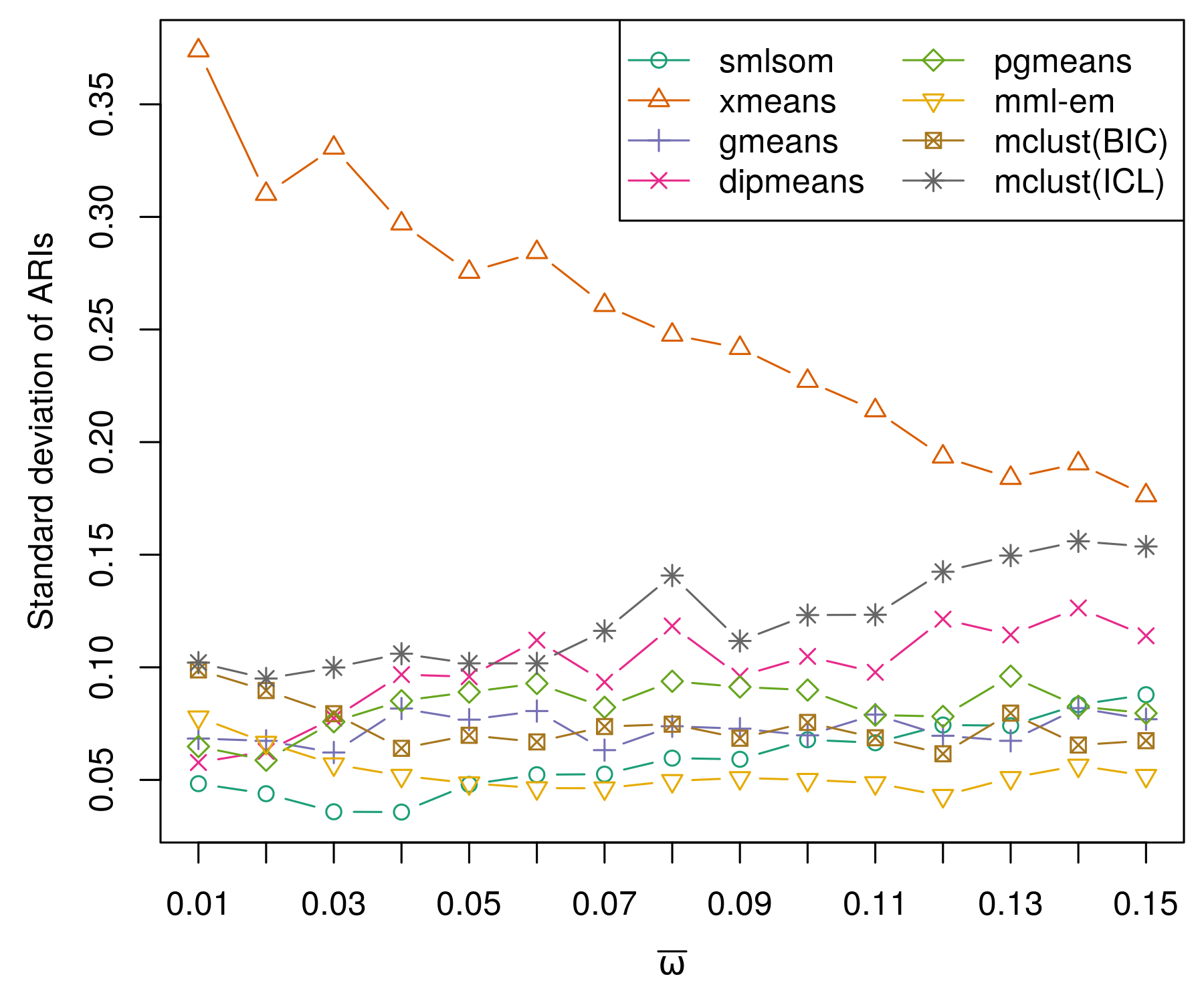} \\
  (a) Average &
  (b) SD
 \end{tabular}
 \caption{Accuracy and stability of
clustering by each method. Calculated scores over
 1000 simulations for each $\bar{\omega}$: (a) Average of ARIs, (b) Standard deviation of ARIs.}
 \label{220212_29Mar17}
\end{figure*}

Fig.~\ref{220212_29Mar17}(a) shows the accuracy and stability of clustering by each method. Note that only ARI results are shown, as NMI had similar results. The result shows that Mclust(BIC), MML-EM, and SMLSOM show the highest accuracy among all methods. In particular, SMLSOM was superior where $\omega$ is small, and MML-EM was superior where $\omega$ is large. In addition, Fig.~\ref{220212_29Mar17}(b) shows that SMLSOM is the most stable of the eight methods when $\omega$ is small, but lacks stability when $\omega$ is large as in Fig.~\ref{fig:selections behavior}(b). MML-EM was stable on average.

G-means did not estimate $M^*$ reasonably, as shown in Fig.~\ref{fig:selections behavior}(a). On the other hand, Fig.~\ref{220212_29Mar17}(a) shows that the evaluation was not that worse. This gap means that although the number of clusters was overestimated, the individual clusters collected samples with specific labels. Thus, the splitting rule for bisecting a cluster is problematic. As pointed out in \cite{feng2007pg}, when $\omega$ is large, k-means is hard-assignment, and the within-cluster sample distribution becomes like a truncated distribution, which is not consistent with a Gaussian distribution. It causes an overestimation of the number of clusters. On the other hand, projecting clusters only in the direction of maximum variance will be worked if the actual clusters are far enough from each other. Otherwise, it is not necessarily a good separation axis to discover two groups.

Although PG-means does not have the same testing problem as G-means above because the test is performed on the entire data, the problem of choosing the projection axis also exists. PG-means recommends generating the projection randomly, but the possibility of finding the appropriate projection axis is low when actual clusters overlap. Since Gaussian distributions with significant overlap look like a single Gaussian distribution, the null hypothesis is often not rejected in the sample after projection, even if $\hat{M}$ is smaller than the true number of clusters. Therefore, the author's recommendation of $N_{\rm prj}=12$ may be insufficient in some cases.
Dip-means performs a unimodality test on the distance distribution of each point to the other points within a cluster. If the true clusters are well-separated, the distribution of distances is bimodal, and Dip-means can work well. However, if the true clusters are close, the distances from any point do not show bimodality.

The difficulty with these statistical testing methods lies in setting appropriate hypotheses, which vary from situation to situation, and in the fact that the sample used for those statistical testings must be one-dimensional.

\begin{figure}[t]
 \centering
 \includegraphics[bb=0 0 432 360, width=9cm]{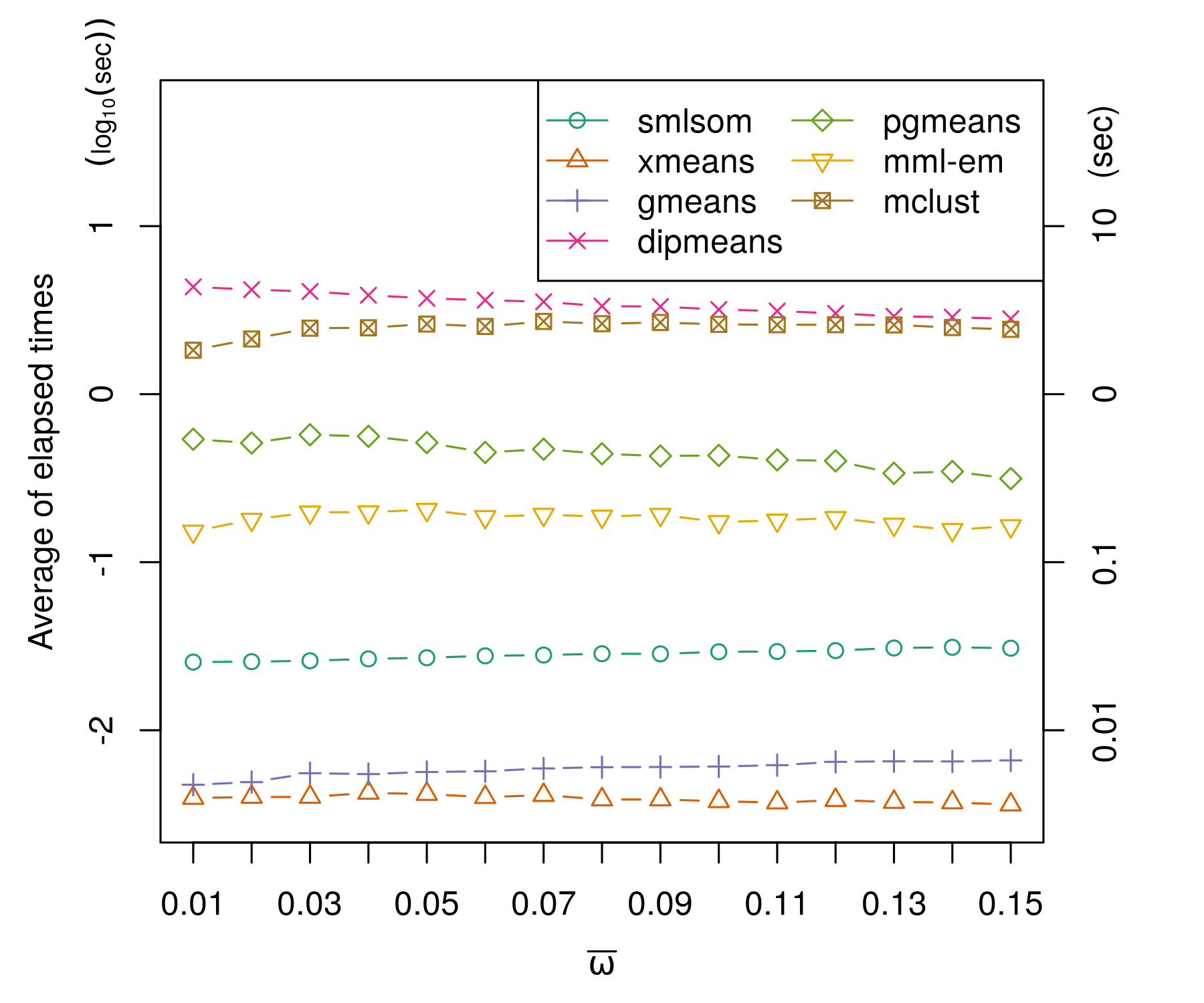}
 \caption{Computational time of each method. The average of
 elapsed time (sec) over 1000 simulations for each $\bar{\omega}$ are shown on a logarithmic scale.}
 \label{005005_30Mar17}
\end{figure}

Unlike the statistical test-based methods described above, X-means, Mclust, MML-EM, and SMLSOM estimate the number of clusters based on the model selection using information criterion. The question is which information criterion to use and what model to use. Fig.~\ref{fig:selections behavior}(a) and Fig.~\ref{220212_29Mar17}(a) shows X-means and Mclust(ICL) did not work well. X-means uses BIC as a decision criterion like Mclust(BIC). However, the Gaussian model X-means uses is too simple and might not represent the sample distribution adequately. Mclust(ICL) was much less accurate than Mclust(BIC). In this case, ICL may not be suitable as a selection criterion.
Mclust(BIC), MML-EM, and SMLSOM answer the above questions to some extent. MML-EM and SMLSOM use selection criteria that consider the selection of the number of clusters, and each produces good estimation results. Mclust uses the widely used standard BIC and works well in this case, although some literature reports a tendency of the BIC to select an excessive number of clusters~\cite{biernacki2000assessing}.

Fig.~\ref{005005_30Mar17} shows the computational time required by each method. The results reveal that the EM algorithm~(Mclust) and Dip-means were the slowest among the methods. Compared with the EM, MML-EM is approximately $15$ times faster, SMLSOM is approximately $80\sim 100$ times faster, and X-means is, roughly speaking, approximately $700$ times faster, and this is the fastest method among all methods.

X-means and G-means are high-speed methods; however, they almost fail to estimate $M^*$ when the cluster overlap is significant. MML-EM succeeded in avoiding the drawbacks of EM, that is, initial parameter dependence and slow convergence. However, the computational time remains large compared to that of SMLSOM. SMLSOM achieved high performance with a lower computational time compared with the EM-based method. Of course, the selection of $M$ by SMLSOM may vary significantly when the distribution overlap is significant. When the overlap is considerable, there is not always one valid $M$, as mentioned earlier. As shown in Fig.~\ref{220212_29Mar17}(a), the variation in the choice of $M$ does not compromise the consistency of the clustering content. 

The stability of the SMLSOM estimation seems to be due to the ``soft-to-hard'' learning rule of SOM. It is known that the effect of the initial position of centroids for final positions approaches zero as the learning progresses of SOM when the learning parameters follow the conditions~\cite{yin1995distribution}. This property is also verified using Monte Carlo simulations, which show that SOM is insensitive to the choice of initial positions~\cite{cottrell2001statistical}. SMLSOM inherits this advantage of SOM. By learning in a ``soft'' manner, the nodes first gather in the center of a dense region, regardless of their initial positions. This property is expected to stabilize the SMLSOM estimation.

\subsubsection{The determination of $\beta$ in SMLSOM}
\begin{figure}[t]
 \centering
 \includegraphics[bb=0 0 432 360, width=10cm, height=6cm]{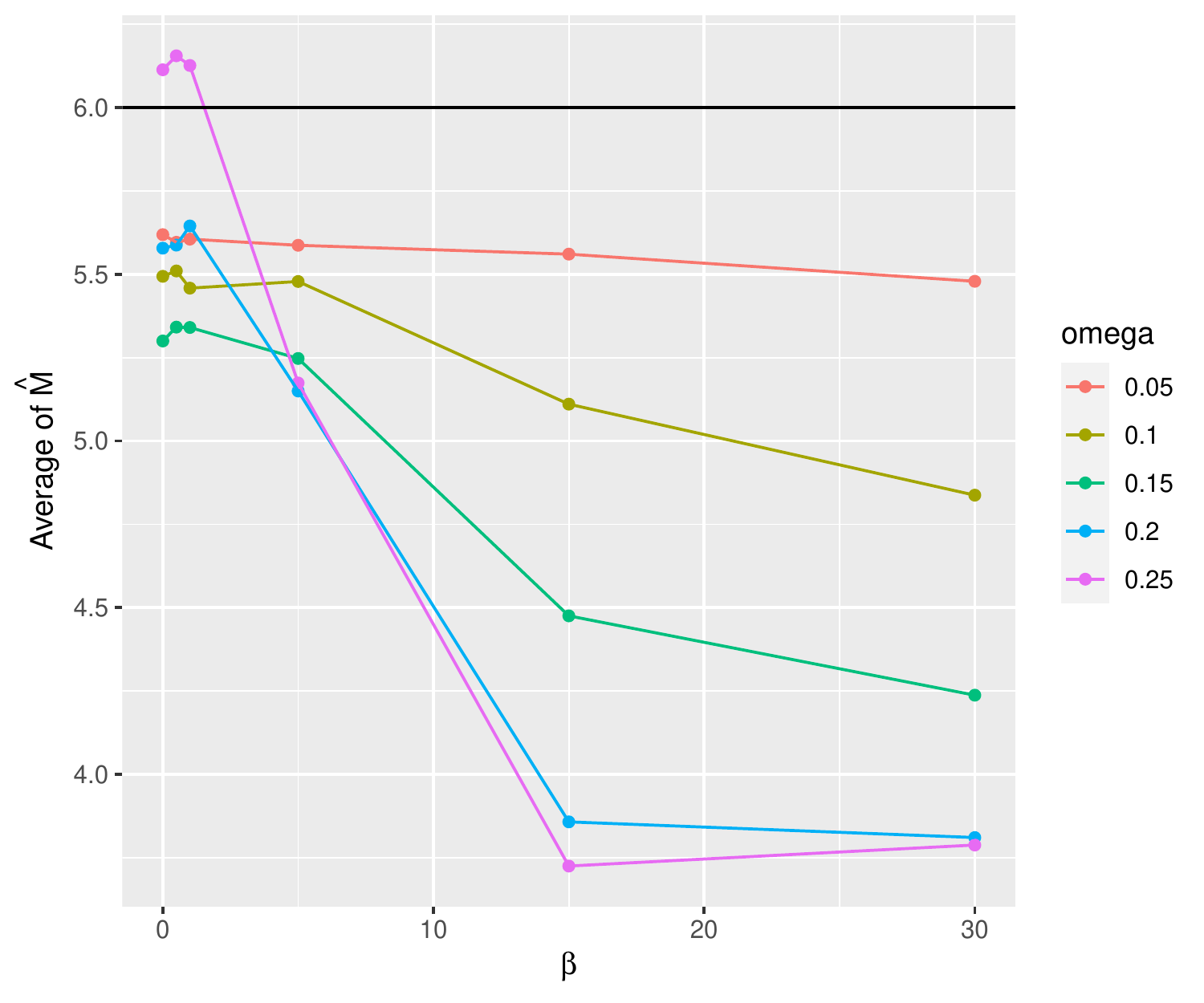}
\caption{Estimation of the number of clusters by SMLSOM with $\beta$ varying.}
 \label{result_beta}
\end{figure}

Finally, the change in SMLSOM estimation results due to $\beta$ is shown in Fig.~\ref{result_beta}. For $\beta$ we tried 0, 0.5, 1, 5, 15, and 30. We generated 20 datasets for each $\omega$ with the same settings as in the previous clause of the experiment.
The larger value of $\beta$, the more difficult the link is to cut. The black horizontal line in the figure is $M^*$. In this experiment, the smaller $\beta$ is, i.e., the easier it is to cut the link, the more accurate and stable the estimation is regardless of $\omega$.

This result shows that if $\omega$ is not considerable, the results do not depend much on the $\beta$ setting. However, if the overlap is significant, it could affect the results.

It is not easy to give a general $\beta$ setting method. Here, we only give practical guidelines based on the author's experience.

The algorithm may terminate without reducing the number of clusters from $M_{\rm max}$. In this case, increasing $M_{\rm max}$, increasing $\beta$, or decreasing $\tau_{\rm max}$ may resolve the problem. The reason for increasing $M_{\rm max}$ is that the number of reasonable clusters may be larger than that. Increasing $\beta$ makes the links harder to cut and the nodes more similar. Almost identical nodes will be removed based on the MDL. The reason why decreasing $\tau_{\rm max}$ is that it may be too adapted to local solutions. Such a solution may have a good likelihood at first glance but yield an unjustified model with too small a variance.
Conversely, the algorithm may terminate with an estimation result that seems too small, such as $M=1$. In this case, the user can try the reverse of the above, but in such cases, the probability distribution model is often not appropriate for the data in the first place. It is necessary to prepare an appropriate model.

We recommend trying several of the above when clustering without pre-knowledge about the data, excluding the extreme results, and then adopting the best MDL result.

\subsection{Continuous data}

We used the MNIST~\cite{lecun1998gradient} datasets for the image clustering experiment. The dataset is a standard handwritten digit dataset containing $28\times 28$ grayscale images, divided into 60,000 training samples and 10,000 test samples.

In MNIST data, each sample is pre-assigned a label from 0 to 9. This experiment aims to understand the latent diversity that labels cannot capture only by searching for the number of clusters without providing information about the labels. Therefore, the purpose of this experiment is not to predict labels. For this reason, we emphasized NMI rather than ARI in this experiment due to the evaluation indexes' natures described in \ref{clause: evaluation}. However, for fairness, the ARI reportings will also be included.

The original data contains the brightness values of each pixel, and the sample dimension is $28\times 28 = 784$, which is a high dimension. To obtain practical features for classification, we performed feature extraction using the HOG descriptor~\cite{dalal2005histograms}. A 324-dimensional feature vector was obtained by adopting nine orientation bins using a $6\times 6=36$ block division. Furthermore, we performed PCA for the HOG feature vectors to reduce the number of features. This dimensionality reduction is called in the HOG-PCA method~\cite{savakis2014efficient}. We adopted 48 components with a cumulative contribution rate of approximately 80\%. 

For comparison, we applied the same methods as in the previous section. The experimental procedure was as follows:  X-means, G-means, and PG-means started with $M_{\rm min}=2$; Mclust ran from $M_{\rm min}=2$ to $M_{\rm max}=100$ each; MML-EM and SMLSOM started with $M_{\rm max}=64$; The map of SMLSOM was $8\times 8$, which was initialized by the method of \ref{Init PCA}. Note that Dip-means when $M_{\rm min}=2,\,3$ did not increase from the initial values, so we set $M_{\rm min}=4$. SMLSOM when $\tau_{\rm max}=n$ did not decrease from $M_{\rm max}$, so we set $\tau_{\rm max}=10000$. Other parameters were the same setting as \ref{clause: parameter_setting}. Each method was run 20 times with training samples.
\begin{figure*}[t]
 \centering
 \hspace*{-1cm}
 \begin{tabular}{cc}
  \includegraphics[bb=0 0 432 360, width=7cm]{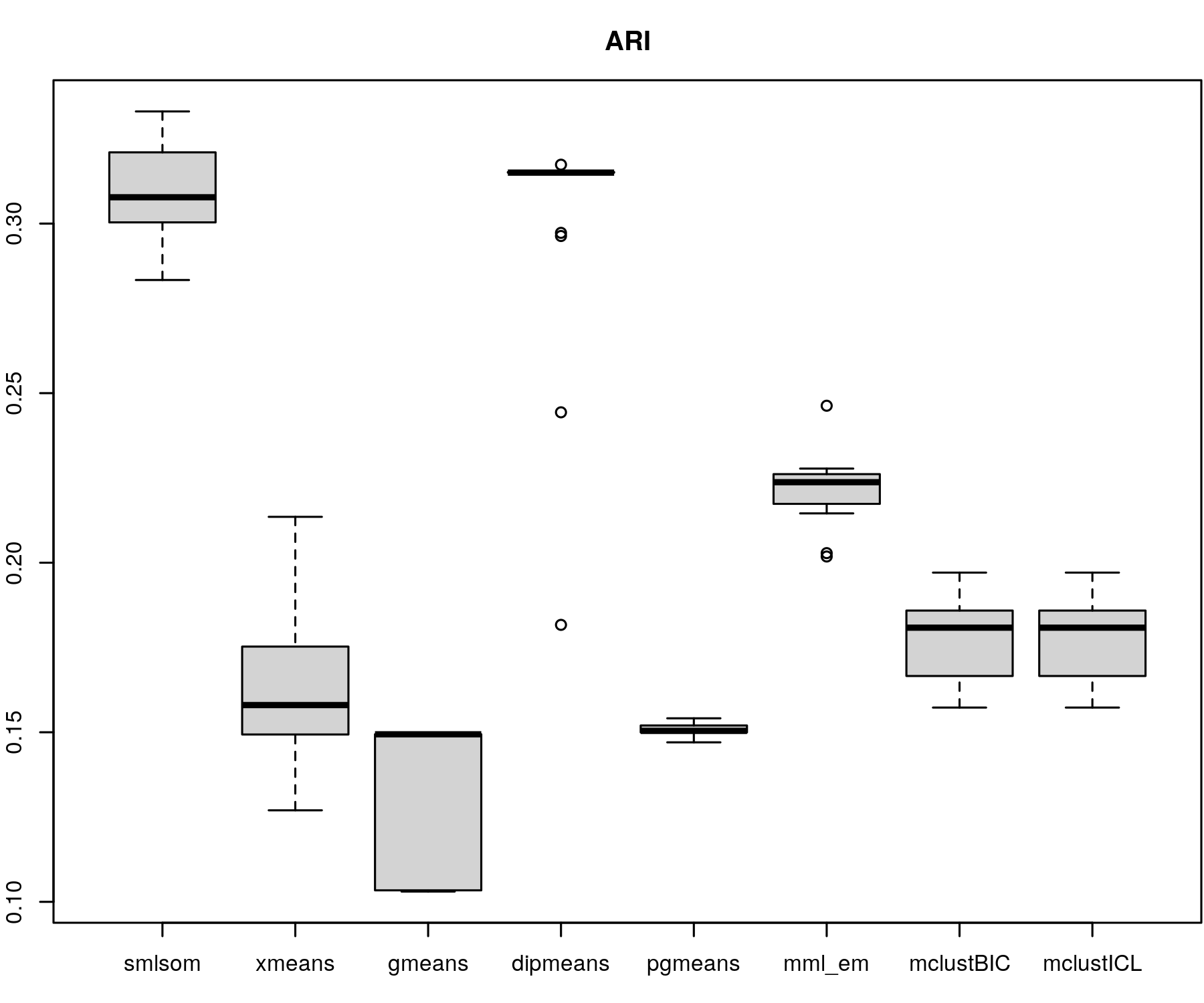} &
  \includegraphics[bb=0 0 432 360, width=7cm]{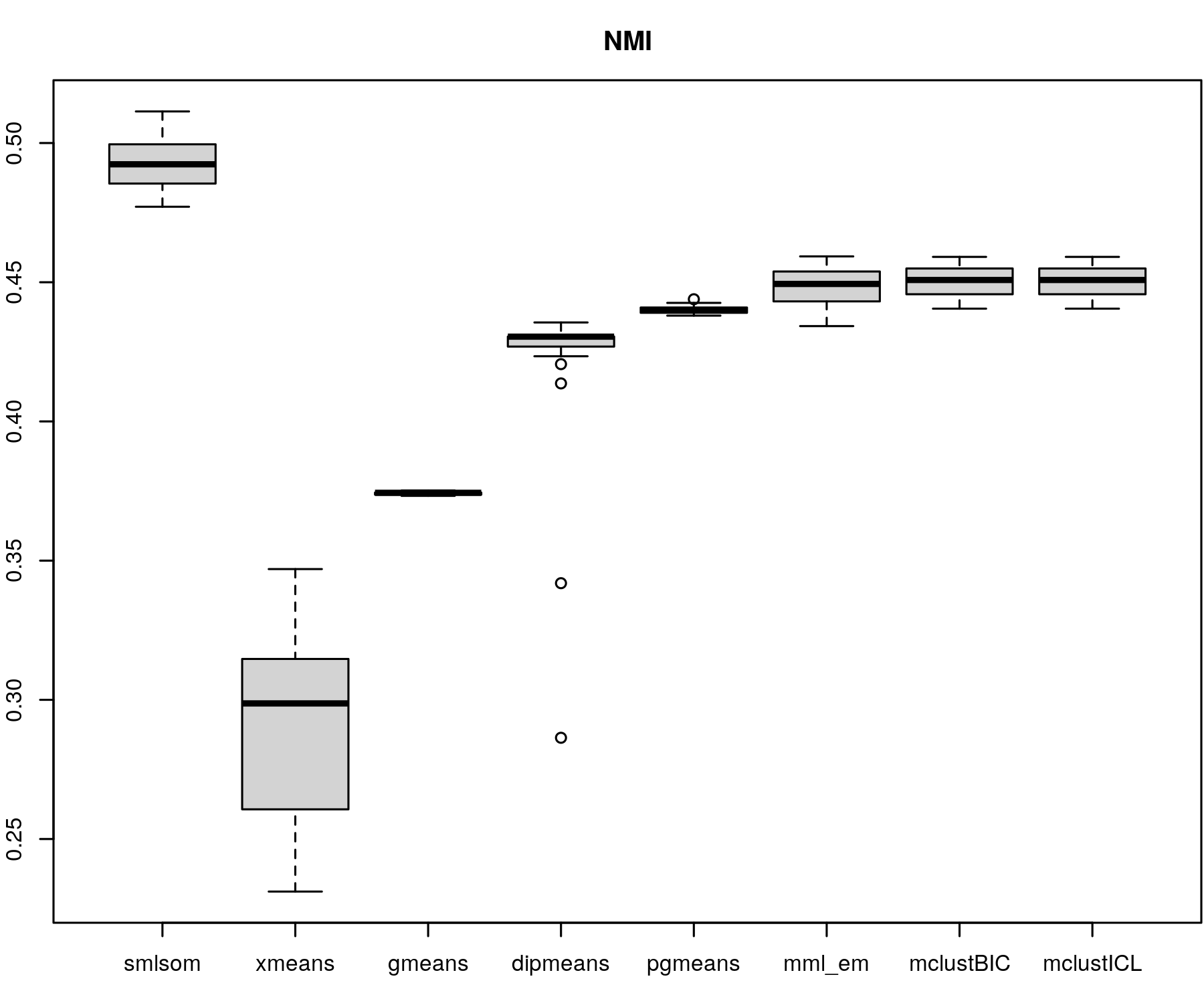} \\
  (a) ARI evaluations for 20 trials &
  (b) NMI evaluations for 20 trials
 \end{tabular}
 \caption{Evaluations for 20 trials in training samples: (a) ARI, (b) NMI.}
 \label{evaluation_training}
\end{figure*}
\begin{table*}[b]
  \caption{Averages of $\hat{M}$ with 20 trials. Standard deviation in the parenthesis.}
  \label{evaluation_training_nd}
  \centering
  \footnotesize
  \begin{tabular}{m{3.5em}m{2.8em}m{2.8em}b{3.5em}b{3.3em}b{3.2em}b{4.6em}b{4.7em}} \hline
    SMLSOM & Xmeans & Gmeans & Dipmeans & PGmeans & MMLEM & Mclust(BIC) & Mclust(ICL)\\ \hline
    36.4 & 100 & 100 & 8.1 & 100 & 60.4 & 93.1 & 93.1\\
    (2.58) & (0.00) & (0.00) & (1.17) & (0.00) & (1.61) & (4.96) & (4.96)\\ \hline
  \end{tabular}
\end{table*}

\begin{table*}[t]
  \caption{Averages of elapsed time (sec) with 20 trials. Standard deviation in the parenthesis.}
  \label{evaluation_training_elapsed}
  \centering
  \footnotesize
  \begin{tabular}{ccccccc} \hline
    SMLSOM & Xmeans & Gmeans & Dipmeans & PGmeans & MMLEM & Mclust\\ \hline
    309.7 & 1.0 & 1.5 & 798.6 & 7900.6 & 659.7 & 3034.4\\
    (28.50) & (0.27) & (0.21) & (143.71) & (56.79) & (46.91) & (476.58)\\ \hline
  \end{tabular}
\end{table*}

Fig.~\ref{evaluation_training} shows evaluations of each method in training samples. The figure shows that SMLSOM was superior to other methods in both evaluation indices. Note that X-means, G-means, Dip-means, and PG-means were evaluated differently depending on the index. This difference is due to their estimated number of clusters and their clustering purity. Table~\ref{evaluation_training_nd} shows the average and standard deviation of $\hat{M}$ with 20 trials. The table shows that X-means, G-means, and PG-means always estimate $\hat{M}=100$, which is equal to $M_{\rm max}$. However, Dip-means estimated the lowest $\hat{M}$ among all methods. ARI highly evaluates the clustering closer to the actual number of labels. On the other hand, a higher NMI with the same number of clusters means that each cluster collects more of a particular label. Table~\ref{evaluation_training_nd} also shows that Mclust(BIC) and Mclust(ICL) selected the same number of clusters. In this case, the results of both methods were not different.

Table~\ref{evaluation_training_elapsed} shows the average elapsed time in training samples. The table shows that although the proposed method is slower than X-means and G-means, the computation time is practical.

Among X-means, G-means, Dip-means, and PG-means, Dip-means stands out in the ARI evaluation because it estimated the most conservative number of clusters. However, PG-means, which estimates $\hat{M}=100$, has a higher evaluation in NMI. It is difficult to judge the superiority of the methods based on the results of the training sample only since these may be over- or underestimates. Therefore, we evaluated the test samples using the clustering results with the best ARI and NMI of the 20 trials in the training sample, respectively. The classification of the test sample into clusters followed the method of each algorithm: X-means, G-means, and Dip-means were using Euclidean distance; PG-means, MML-EM, and Mclust classified samples into clusters with the maximum posterior probability; SMLSOM classified with the maximum likelihood.

\begin{table}[t]
  \caption{Evaluation of best results}
  \label{MNIST_evaluation_test}
  \centering
  \footnotesize
  \begin{tabular}{lrrrcrrr} \hline
    & \multicolumn{3}{c}{ARI select} && \multicolumn{3}{c}{NMI select} \\ \cline{2-4} \cline{6-8}
    & $\hat{M}^*$ & Training ARI & Test ARI && $\hat{M}^*$ & Training NMI & Test NMI\\ \hline
    SMLSOM & 34 & \bf{0.333} & \bf{0.350} && 34 & \bf{0.511} & \bf{0.524}\\
    X-means & 100 & 0.124 & 0.038 && 100 & 0.347 & 0.117\\
    G-means & 100 & 0.150 & 0.037 && 100 & 0.375 & 0.109\\
    Dip-means & 7 & 0.317 & 0.061 && 9 & 0.436 & 0.056\\
    PG-means & 100 & 0.154 & 0.158 && 100 & 0.444 & 0.450\\
    MML-EM & 56 & 0.246 & 0.257 && 60 & 0.459 & 0.465\\
    Mclust(BIC) & 84 & 0.197 & 0.211 && 84 & 0.459 & 0.469\\
    Mclust(ICL) & 84 & 0.197 & 0.211 && 84 & 0.459 & 0.469\\
    \hline
  \end{tabular}
\end{table}

The results are in Table~\ref{MNIST_evaluation_test}. The table lists the estimated number of clusters, training evaluation, and test evaluation for the ARI and NMI best clustering results, respectively. From the table, it can be seen that the SMLSOM results are the best for all indicators. On the other hand, for X-means, G-means, and Dip-means, the test accuracy was lower than the training accuracy for all indicators. In particular, Dip-means had a large drop in evaluation. Contrarily, PG-means, MML-EM, Mclust, and SMLSOM did not deteriorate the test evaluation.

\begin{figure}[t]
  \centering
  \includegraphics[bb=0 0 576 576, width=10cm, height=10cm]{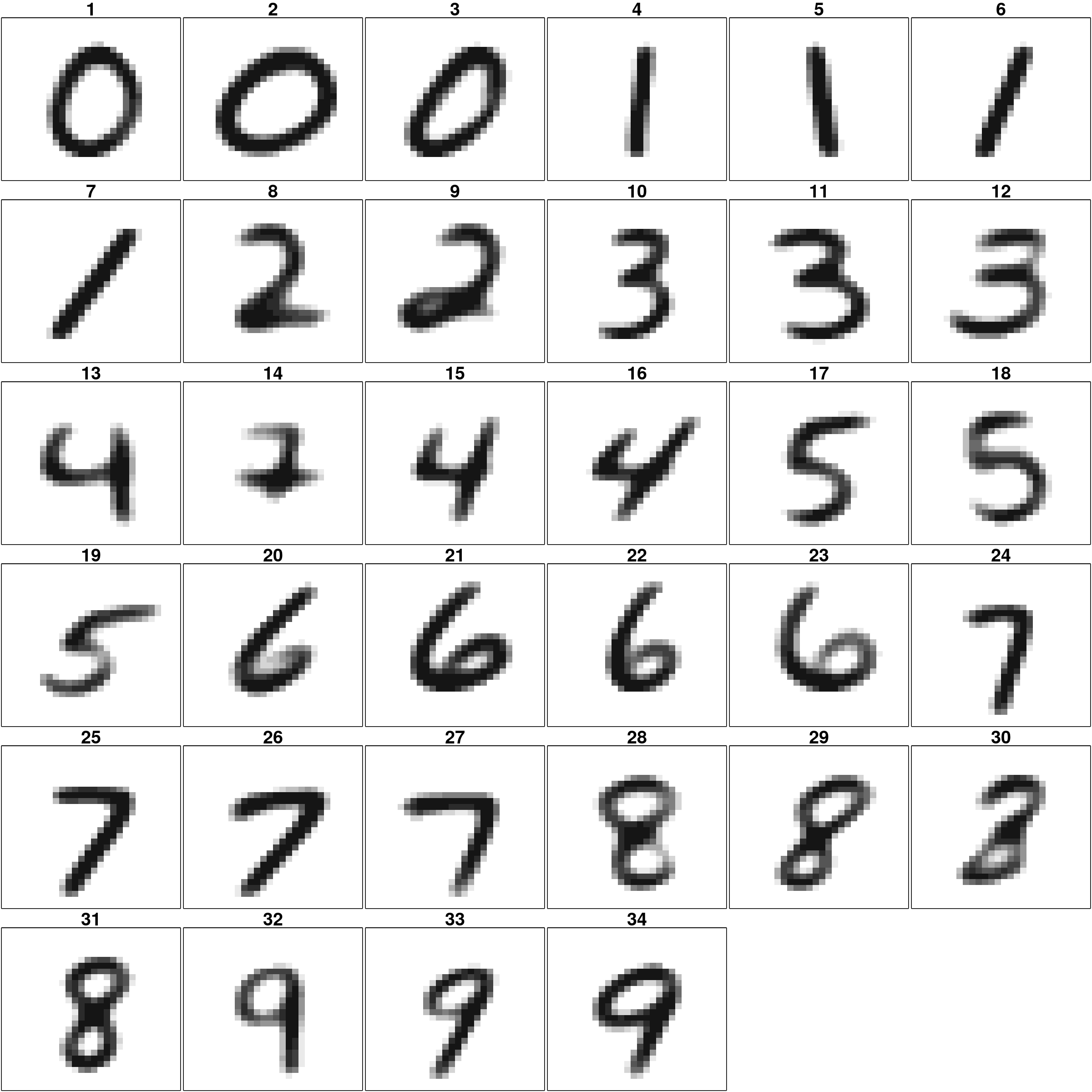}
  \caption{Median images from each of the 34 clusters.}
  \label{MNIST_plot}
\end{figure}

Finally, the most highly evaluated SMLSOM clustering result was examined to see what samples each cluster collected. The best NMI result was used here, although both were similar.

Fig.~\ref{MNIST_plot} shows the median image of images belonging to each cluster in the NMI best result of SMLSOM trials. For some clusters, multiple digits are mixed, but in the other many clusters, digits are well separated. The clustering result shows that the characteristics differ significantly, even if the numbers are the same. The result also indicates that digits are classified by the character shape difference, such as inclination, width, and a loop of the number 2.

This experiment shows that SMLSOM can classify different characteristics, even with the same number of handwritten characters. However, in some clusters, SMLSOM cannot organize the images well. For example, Cluster 14 mixes digits 2, 4, and 7. To separate such characters, it is necessary to explore more expressive features or models.

As described above, the proposed method achieved higher performance in lower computation time than other methods. We also showed that the proposed method could provide a reasonable number of clusters, roughly corresponding to each digit. Note that for high-dimensional data such as MNIST data, feature extraction to reduce dimensionality will be necessary so that algorithms can work well.

\subsection{Count data}

\begin{figure}[t]
 \centering
 \includegraphics[bb=0 0 432 216, width=13cm]{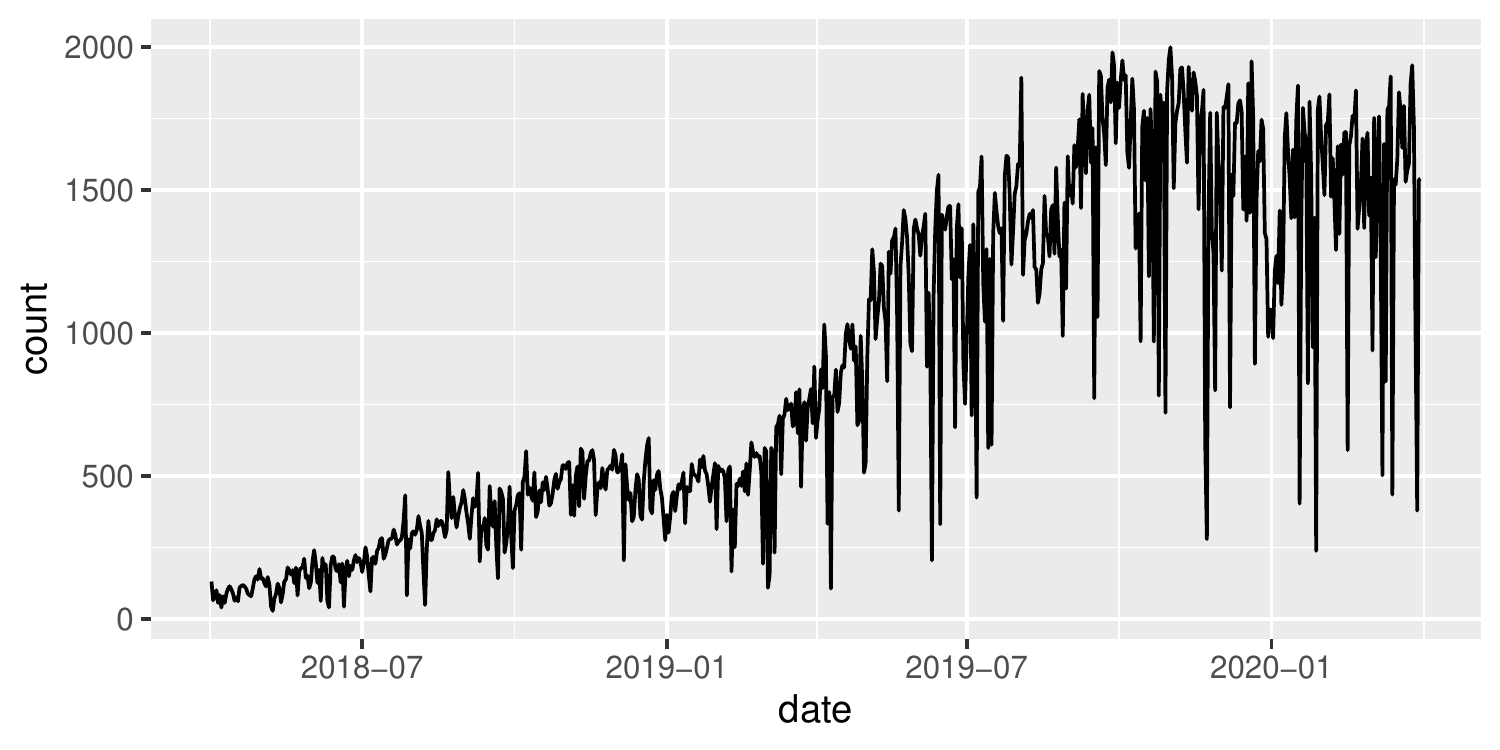}
 \caption{Number of uses per day for all share cycle stations in Chiba city from April 2018 to March 2020.}
 \label{sharecycle_daily_count}
\end{figure}

In this experiment, we used the open data of the share-cycle system in Chiba City, Japan, from 2018 to 2020\footnote{Chiba City Share Cycle Open Data. Available at: \url{https://www.city.chiba.jp/sogoseisaku/miraitoshi/tokku/share-cycle_opendata.html}, accessed 2021/11.}. Chiba City has set up bicycle stations throughout the city. Users can rent bicycles using IC cards or online reservations, and return them where they want to go. Chiba City published data from April 2018 to March 2020. The data specify the number of uses (rentals and returns) by location per hour. Fig.~\ref{sharecycle_daily_count} shows the daily usage numbers from April 2018 to March 2020.

The number of bicycles at each station fluctuates depending on the status of the rentals and returns. Therefore, excessive lending will cause a shortage of inventory, and excessive returns will result in a lack of space for park bicycles. Consequently, it is necessary to coordinate operations to bring bikes from other stations or to move stocks to others. For this operation, we focused on lending and returning at each station and analyzed the situations under which the number of bicycles lent and returned would be uneven.

\begin{figure}[t]
 \footnotesize
 \hspace{-2cm}
 \begin{tabular}{cc}
  \begin{minipage}[t]{8cm}
   \includegraphics[bb=0 0 432 360, width=8cm]{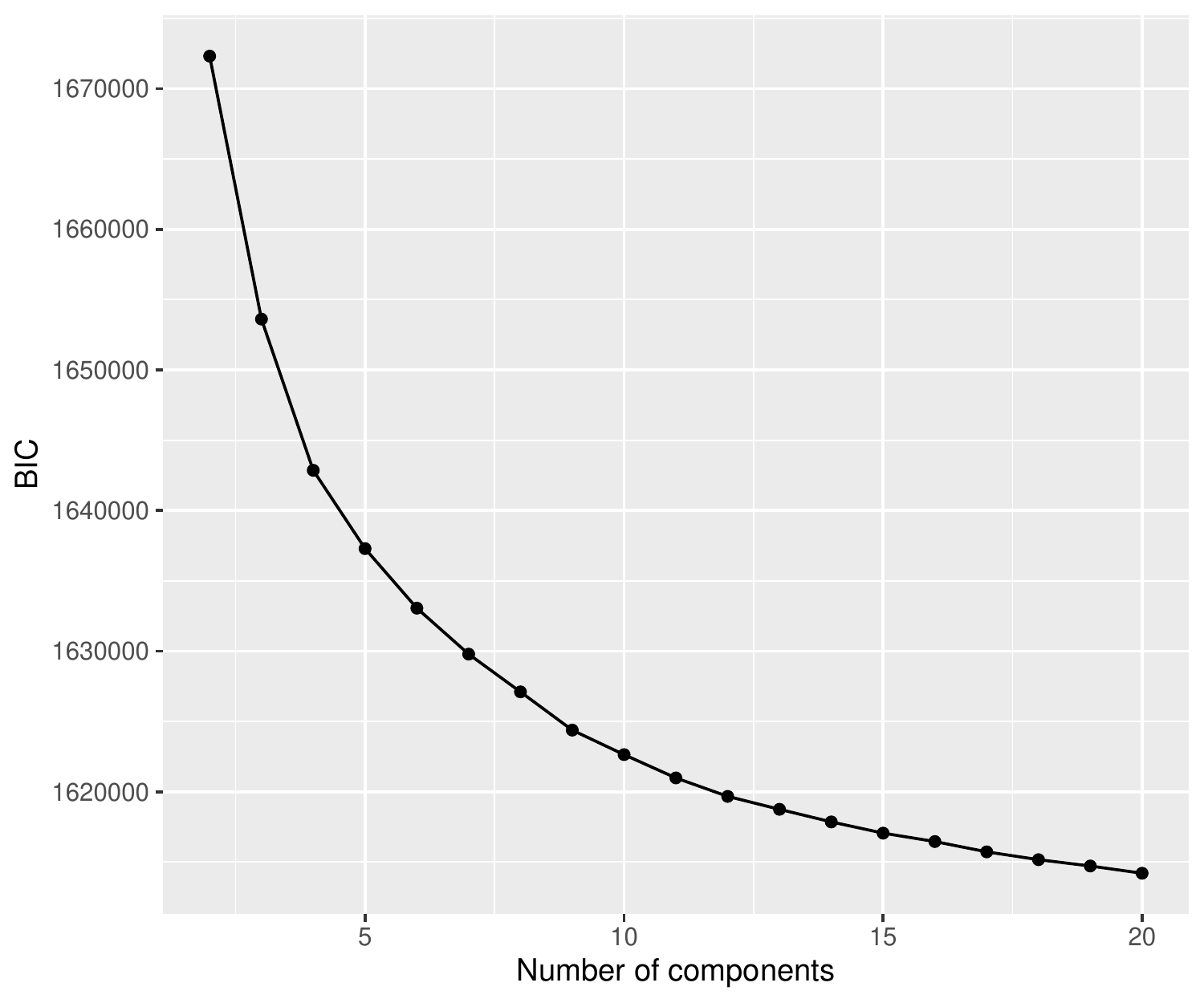}
  \end{minipage} &
  \begin{minipage}[t]{8cm}
   \includegraphics[bb=0 0 432 360, width=8cm]{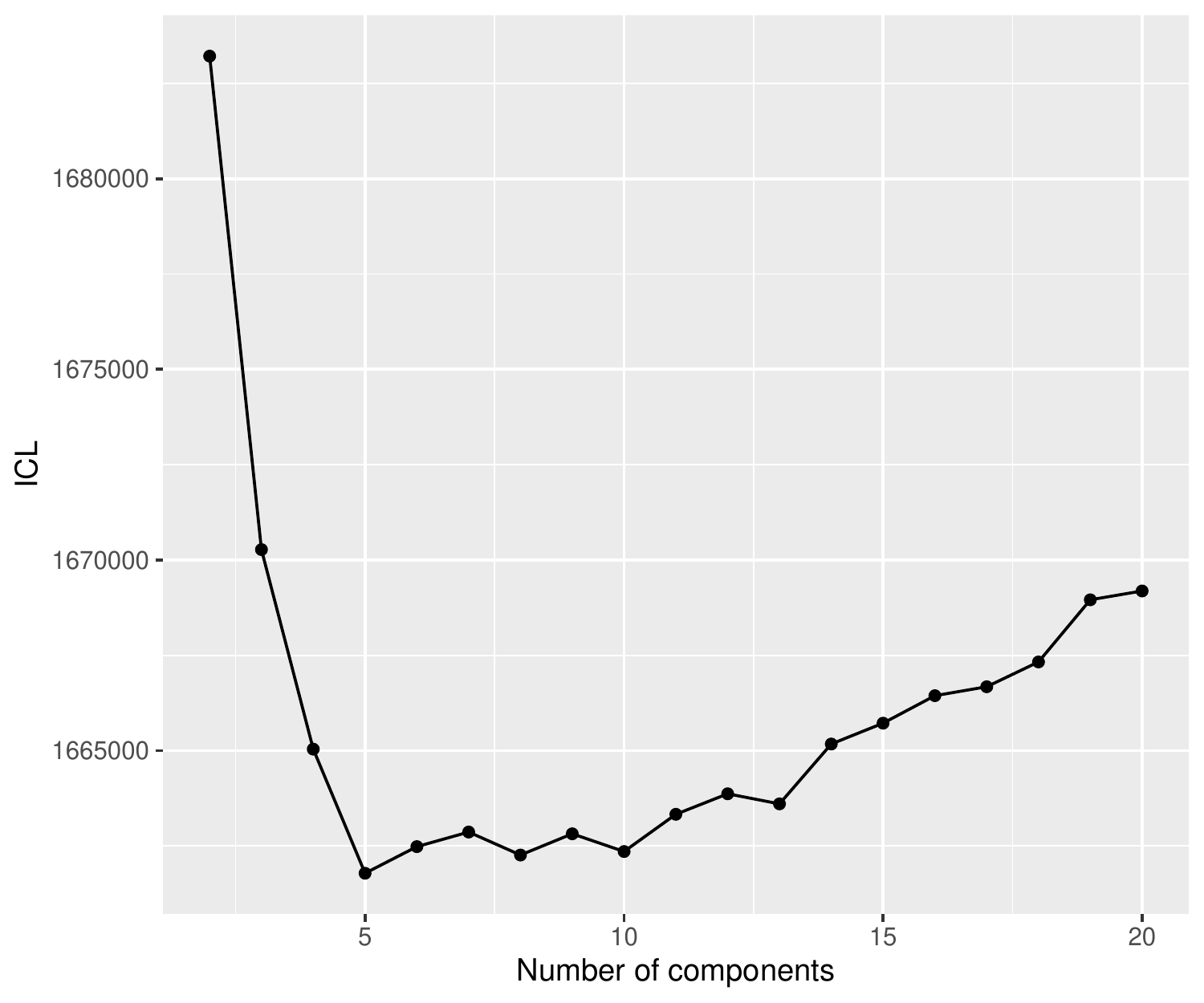}
  \end{minipage} \\
  (a) BIC & (b) ICL
 \end{tabular}
 \caption{Evaluation of the results by EM (multinomial): (a) BIC evaluation, and (b) ICL evaluation.}
 \label{sharecycle_mixtools}
\end{figure}

We used the last one-year data from April 2019 to March 2020 because the number of uses was still low in 2018 when they launched the service. For pre-processing the data, we selected the usage time and target stations. Usage time is the time from borrowing to return. We excluded more extended periods of use because return timing varied greatly. Approximately 7\% of the total usages were more than 60 min, and approximately 5\% of the total usages were more than 90 min; thus, we selected a target usage time of 60 min or less.

Because some stations were newly established during the data period, they have not been used much. We excluded stations with less than 100 days of usage. The number of usages at stations that met this condition was approximately 1\% of the total used for 60 min or less. In addition, we excluded days when users did not use the station.

After pre-processing, we formatted the data to have {\it location} $\times$ {\it date} as a row, and the number of uses per hour for each return and lending as a column. The number of target stations was 240. The sample size was 73,342, and the number of columns was $24 \times 2 = 48$. By clustering the data, we created daily patterns of the usage frequency time series.

A representative model for multivariate count data is a multinomial distribution. We ran SMLSOM and EM for data with multinomial distributions. SMLSOM was run 10 times using $\beta = 15$ and a $5\times 4$ hexagonal lattice with random initial values. EM was run 10 times for each component number from 2 to 20, and the initial value for each component number was randomly changed. SMLSOM selected five clusters, and EM also selected the same number, as shown in Fig.~\ref{sharecycle_mixtools}(b).

\begin{figure*}[t]
 \footnotesize
 \hspace{-2cm}
 \begin{tabular}{cc}
  \begin{minipage}[t]{8cm}
   \includegraphics[bb=0 0 432 432, width=8cm]{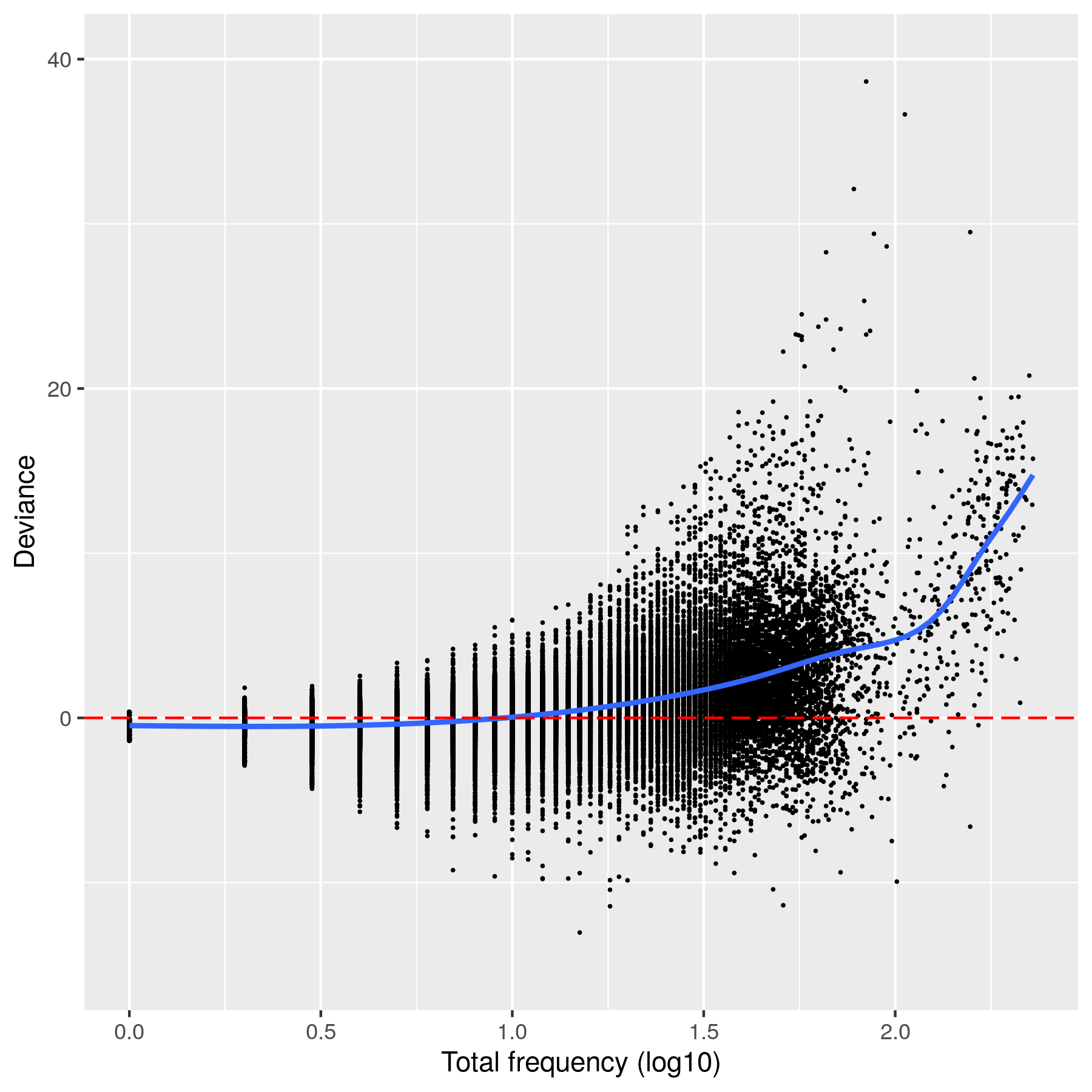}
  \end{minipage} &
  \begin{minipage}[t]{8cm}
   \includegraphics[bb=0 0 432 432, width=8cm]{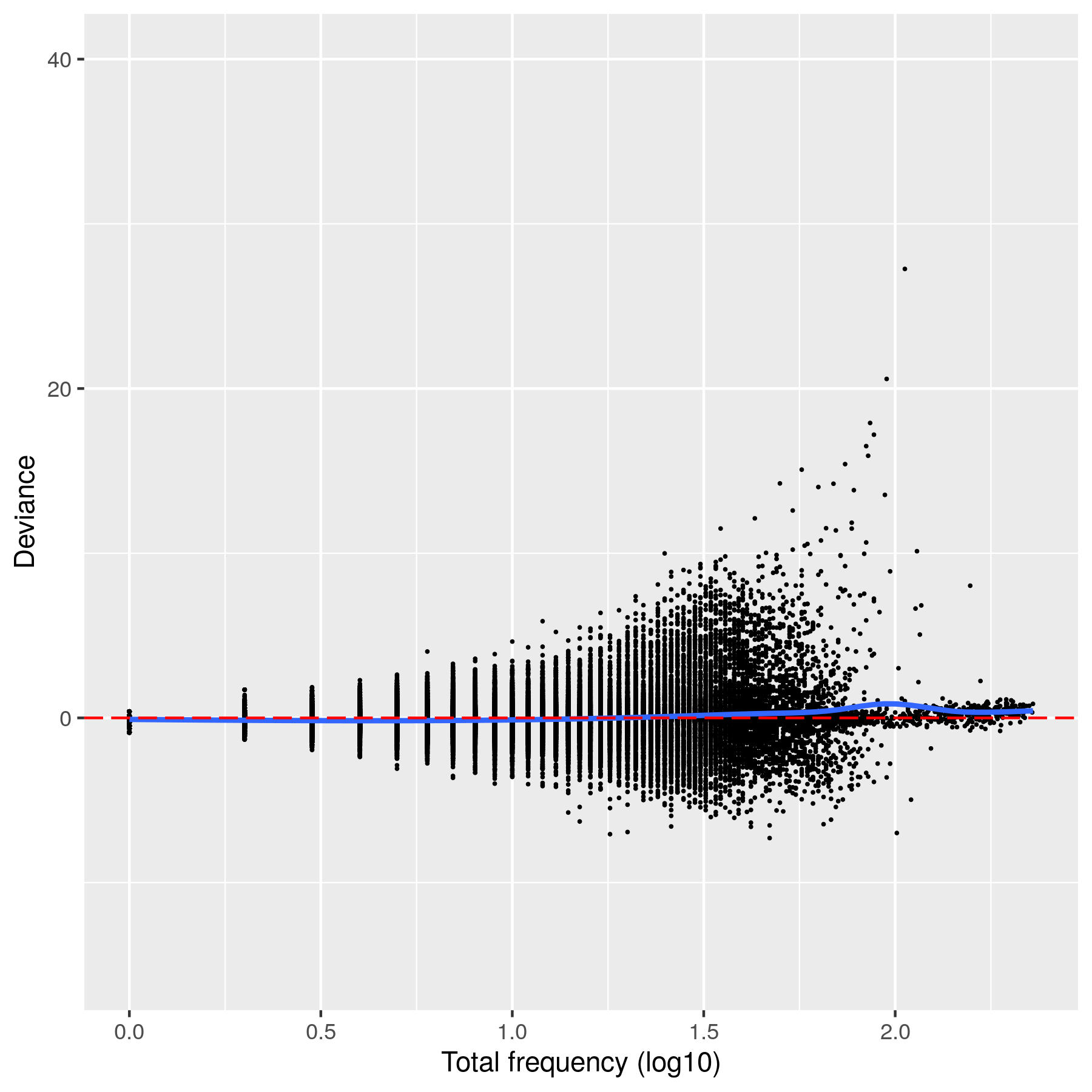}
  \end{minipage} \\
  (a) & (b)
 \end{tabular}
 \caption{Total frequency and deviation between two models: (a) Deviation between EM and SMLSOM. (b) Deviation between EM and SMLSOM+EM. The blue line represents a smoothing curve using cubic splines.}
 \label{deviance}
\end{figure*}

To compare the two clustering results with the same likelihood criterion, we evaluated the EM result using the MDL criterion in Eq.~(\ref{cmdl}). Note that when calculating $-\log L_C(\hat{\bmT})$, $\hat{m}_i$ is assumed to be the cluster with the largest posterior probability. The MDL evaluation of the SMLSOM result was 1,695,888, while the MDL evaluation of the EM result was 1,678,087, indicating that the EM result was slightly better. To investigate the fit of the two estimated models to the data, we defined the deviance of the two models for each sample as follows:

$$\log f(\bm{x}_i \mid \hat{\bm{\theta}}'_{\hat{m}'_i}) - \log f(\bm{x}_i \mid \hat{\bm{\theta}}_{\hat{m}_i}),$$
where $\hat{\bm{\theta}}'_{\hat{m}'_i}$ and $\hat{\bm{\theta}}_{\hat{m}_i}$ are the estimation results of EM and SMLSOM, respectively. Therefore, the deviation is negative if the likelihood is higher for SMLSOM and, conversely, it is positive if the likelihood is higher for EM.

The relationship between this quantity and the total frequency of each sample, $n_i=\sum_{j=1}^p x_{ij}$, is shown in Fig.~\ref{deviance}(a). The figure shows that SMLSOM fits well in terms of likelihood, where the total frequency is small, and EM fits well where the total frequency is large. This tendency can be attributed to the following reasons.

\begin{figure*}[t]
 \footnotesize
 \hspace{-1cm}
 \begin{tabular}{c}
  \includegraphics[bb=0 0 1008 216, width=15cm]{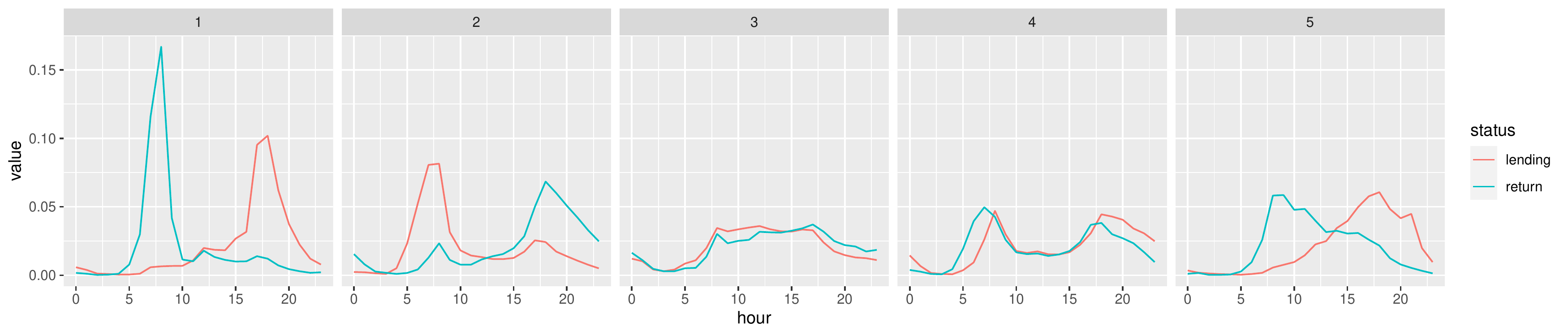}\\
  (a) EM \\
  \includegraphics[bb=0 0 1008 216, width=15cm]{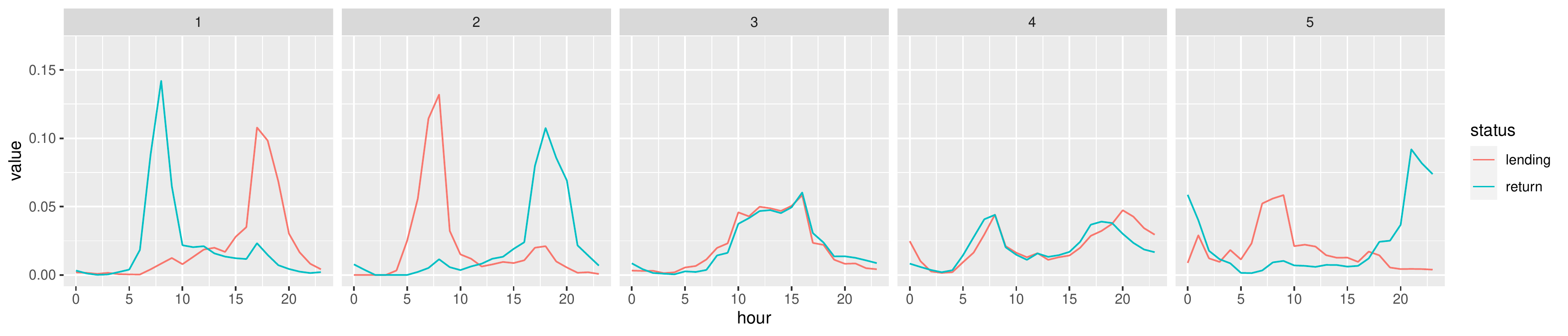}\\
  (b) SMLSOM\\
  \includegraphics[bb=0 0 1008 216, width=15cm]{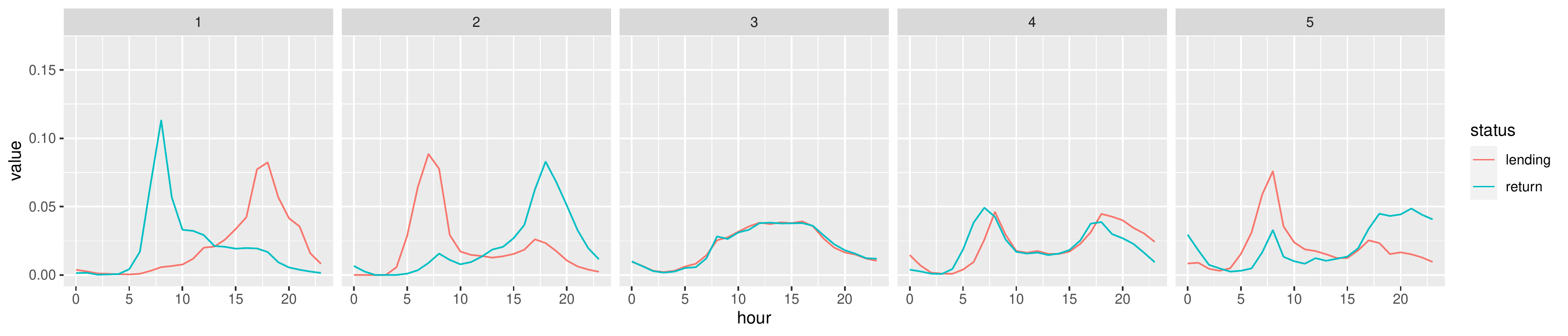}\\
  (c) SMLSOM+EM\\
 \end{tabular}
 \caption{The estimated parameters of the multinomial distribution: (a) EM result, (b) SMLSOM result, (c) SMLSOM+EM result.}
 \label{estimated_timeseries}
\end{figure*}

In the EM algorithm, the average of $x_{ij} / n_i$ weighted by the posterior probability is calculated as an estimate of the parameter $\theta_{mj}$ of the multinomial distribution. Because the $n_i$ term is canceled in the numerator and denominator of the posterior probability calculation, $n_i$ of $x_{ij} / n_i$ directly affects the estimated value. Therefore, the estimation of $\theta_{mj}$ by EM can be strongly affected by the samples with large $n_i$. On the other hand, because SMLSOM updates $\theta_{mj}$ using only the relative frequencies of randomly sampled $\bm{x}_i$, as described in \ref{multinomial model}, both large and small $n_i$ samples are treated equally. The good fit of the EM estimation results at large total frequencies and the relatively poor fit at small frequencies can be due to the difference in the estimation methods described above.

\begin{figure}[t]
 \centering
 \includegraphics[bb=0 0 792 612, clip, viewport=50 0 792 612, width=12cm]{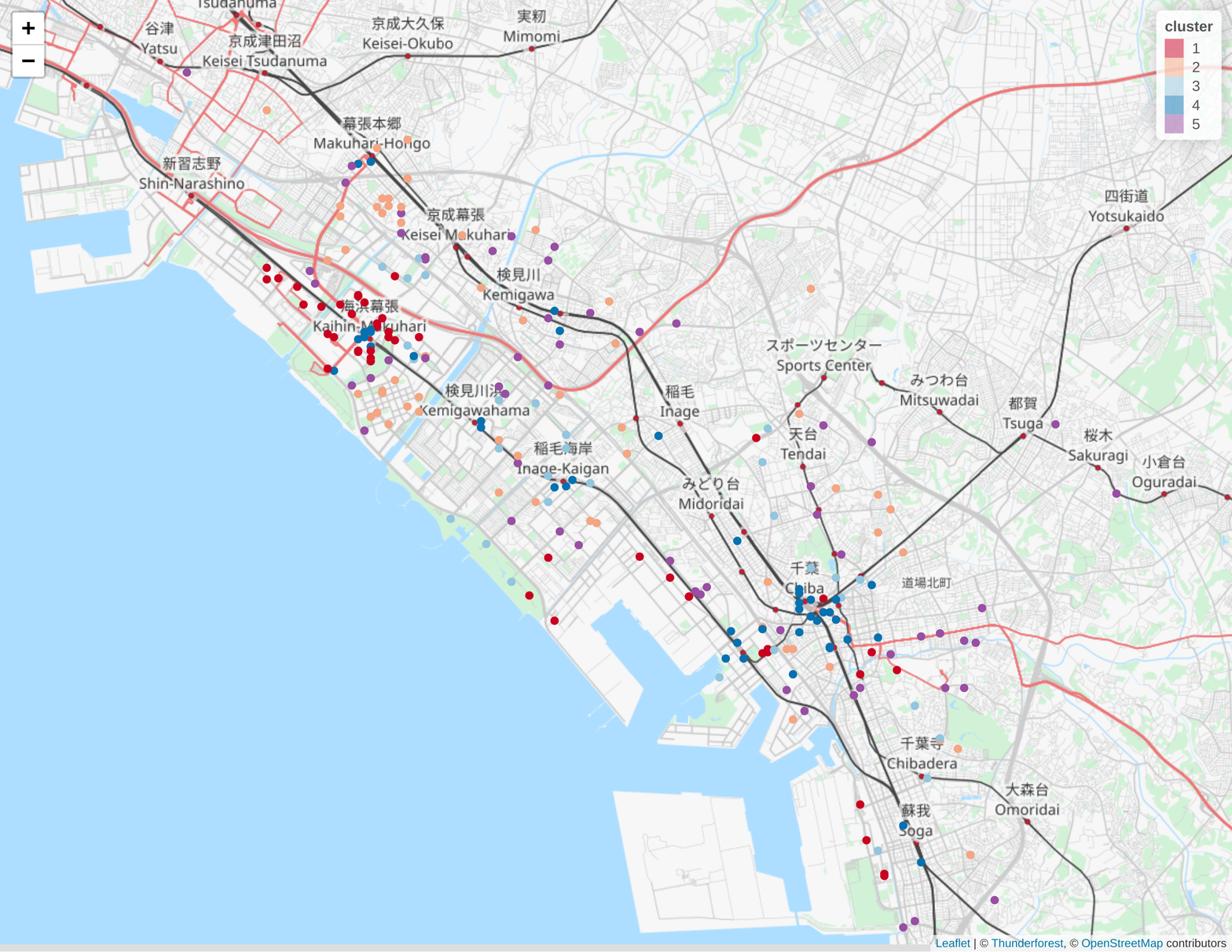}
 \caption{Clustering results on a map using SMLSOM+EM result. As a representative cluster for each station, the cluster with the largest size at that location is chosen.}
 \label{clustering_map}
\end{figure}

Therefore, we attempted to improve the poor fit of SMLSOM in samples with a large total frequency by applying the EM step several times with the SMLSOM estimation results as initial values. The number of EM iterations was evaluated in several trials with the MDL, and we found that approximately 10 iterations were sufficient for this data. Note that the computation time for additional learning by EM has little effect on the overall search time because it is performed only once on the best results of SMLSOM. The resulting MDL of 1,671,663 was slightly better than that of the single EM result (1,678,087). Fig.~\ref{deviance}(b) shows the deviation between this estimation result and the single EM result. It can be seen that the deviation at high frequencies is significantly improved. However, the fitting in the low-frequency part is approximately the same as, or only slightly better than, the single EM result.

Fig.~\ref{estimated_timeseries} shows the estimated values of $\theta_{mj}$ by each method. The two results from single EM and single SMLSOM estimate generally similar patterns, but differ in ``Cluster 5.'' This difference seems to be due to the difference in the estimation methods mentioned above. In fact, in ``Cluster 5'' of Fig.~\ref{estimated_timeseries}(a) the median of $n_i$ is 14, the highest value among the single EM clusters, and in ``Cluster 5'' of Fig.~\ref{estimated_timeseries}(b) the median value is 5, which is the lowest average among the single SMLSOM clusters. In contrast, SMLSOM+EM in Fig.~\ref{estimated_timeseries}(c) takes over the patterns found in the single SMLSOM result, but modifies the time of day where usage increases/decreases and the height of peaks.

Next, the clusters of points are displayed on the map using the results of SMLSOM+EM with the best MDL. Fig.~\ref{clustering_map} shows the representative cluster of each station, which is the cluster with the largest size for the location. Major train stations, such as Kaihin-Makuhari Station and Chiba Station, have ``Cluster 4'' stations with peak usage in the mornings and evenings, but slightly shifting lending and return times. However, a little further away from the train stations, there are ``Cluster 1'' stations, where most returns are in the morning and most rentals are in the evening. This tendency may be due to commuting to work and school.

On the other hand, in residential areas far from the train station, ``Cluster 2'' points are distributed with more returns in the evening and more rentals in the morning. This pattern may be due to commuting, shopping, or leisure activities during the day, and returning home in the evening. However, in some places in the same residential area, there also are ``Cluster 5'' stations where returns continue into the late hours.

As for ``Cluster 3,'' the map does not show a clear rule of distribution; however, from our investigation, it seems that many of the stations are located near public facilities such as universities, libraries, community centers, and parks.

In the actual operation to coordinate bicycle numbers, it is necessary to predict the gap between returns and rentals in advance at the time of day. Because there is a significant time difference between the peaks of returns and rentals in ``Clusters 1 and 2,'' it is easy to predict when the gap is likely to occur. In ``Cluster 3,'' returns and rentals are generally balanced, so that there is little need for adjustment. On the other hand, in ``Cluster 4,'' the pattern of returns and rentals is similar, but the phase is slightly different, so that the gap may appear and disappear in a short period. Thus, a more accurate prediction is required. As with ``Cluster 1'' and ``Cluster 2,'' there is a significant time difference between the return and lending peaks in ``Cluster 5,'' so that predicting the gap itself is easy, but there may be operational issues such as whether there are coordinators who can handle excessive late-night returns. In such a case, it is necessary to forecast the gap by considering the operational constraints.

As described above, the proposed method gave the same number of clusters as EM, but the computation time to find $\hat{M}$ was approximately eight times faster for the proposed method than for EM. We also showed that although the proposed method gave a rougher estimation than EM, additional learning can give more reasonable clusters in the sense of MDL, although it is only slightly better than the EM result.

\section{Disccusion}
\label{disccusion}
As seen in the example of the count data experiment, the SMLSOM may provide rough estimates. This roughness is because this method aims to find appropriate patterns, and the accuracy of the parameter estimates is sufficient to contribute to pattern discovery, and not to make exact estimates.

Our method determines the sample classification using maximum likelihood, but the parameters themselves are estimated using the method of moments instead of the maximum likelihood estimation method, unlike EM and other methods.  This is because there are some distributions for which it is difficult to estimate the parameters using maximum likelihood estimation. For example, the negative binomial distribution has two parameters: the number of trials until the experiment succeeds ($r$), and the probability of success ($p$); however, the maximum likelihood estimator of $r$ cannot be obtained in closed form. On the other hand, according to the method of moments, the estimation of $r$ can be obtained by a simple calculation.

Our method allows data analysts to find a rough idea of the potential patterns in the data without taking too much time. We believe that this will contribute to an understanding of the data in the early stages of data analysis. However, it may be necessary to improve the accuracy of parameter estimation. For example, as in the instance of share cycle data, it may be helpful to use the parameters obtained by SMLSOM as initial values to improve the estimation accuracy by using a more rigorous method such as the EM algorithm.

In this study, we treated all input variables as valuable for clustering. Therefore, this study does not consider high-dimensional data. There may be many unnecessary variables for clustering in high-dimensional data, and clustering may not be successful if such variables become noisy. For example, the MNIST data used in Section~\ref{sec:experiment} are also grayscale data with 784 dimensions. In this experiment, we used feature extraction and dimension reduction to reduce the number of dimensions. When applying the proposed method to high-dimensional data, as in the MNIST dataset, it is necessary to perform appropriate feature extraction in advance.

\section{Conclusion}
\label{sec:conclusion}
In this paper, we proposed a greedy algorithm called SMLSOM to select the number of clusters $M$. In SMLSOM, clusters are constructed using the SOM learning rule. The algorithm then updates the graph structure that connects the probability distribution model to a node. We showed that the dependence on the initial value can be reduced, and that a model appropriately chosen can be compared to the existing method as long as valuable features are given. In addition, the proposed method is applicable to any probability distribution model as long as the distribution function can be calculated by the method of moments. For data where introducing a probability distribution is beneficial, such as count data, we also showed that the proposed method has a lower computational cost than applicable methods such as the EM algorithm.

\section*{Acknowledgement}
This work was supported by JSPS KAKENHI Grant Number JP21H04600 and JST SPRING Grant Number JPMJSP2146.

\appendix

\section{Initialization}
\label{Init PCA}

Let $M=P\times Q$; the initial reference vector of node $m$,
$\bm{\mu}_m^{\rm init}$, is calculated as
\begin{equation}
 \bm{\mu}_m^{\rm init} = \bar{\bm{X}} + A_1(m) \sqrt{\lambda _1}\bm{z}_1 + A_2(m) \sqrt{\lambda _2}\bm{z}_2,
  \label{eq:linear init}
\end{equation}
where $\bar{\bm{X}}=\frac{1}{n}\sum _{i=1}^n \bm{x}_i$, and
$\lambda _1,\,\lambda _2,\,\bm{z}_1,\,\bm{z}_2$ are the first and second largest eigenvalues and corresponding eigenvectors of $\bm{X}^t\bm{X}$, respectively. Further, $A_1(m),\,A_2(m)$ constitute a sequence of numbers from $-2$ to $2$
with a common difference; they are given by
\begin{eqnarray}
 A_1(m) &=& -2 + \{(m-1) {\rm \  mod\ } P\} \frac{4}{P-1},  \\
 A_2(m) &=& -2 + \{\lfloor (m-1)/P \rfloor \} \frac{4}{Q-1}.
\end{eqnarray}

\section{Gaussian model}
\label{gaussian model}
Suppose $\bm{x} = (x_1,\,x_2,\,\ldots ,\,x_p)^t$ follows a
multivariate Gaussian distribution. The probability density function is given by
\begin{eqnarray}
 \lefteqn{f(\bm{x}\mid \bm{\mu},\,\bm{\Sigma})=} \nonumber \\
  &&
  \frac{1}{(2\pi)^{p/2}|\bm{\Sigma}|^{1/2}}\exp
  \left\{-\frac{1}{2}(\bm{x}-\bm{\mu})^t\bm{\Sigma} ^{-1}(\bm{x}-\bm{\mu})\right\}
\end{eqnarray}
where $\bm{\mu}$ is the mean vector and $\bm{\Sigma}$ is the covariance matrix.

Let $\bm{z}$ be the empirical first-order moments, $\bm{Z}$ be the empirical second-order moments, updated using the rule in (\ref{181819_16Jul17}) follows:
\begin{eqnarray}
 \Delta \bm{z} & = & \alpha(\tau)(\bm{x}-\bm{z}),\label{1st_moment_update} \\
 \Delta \bm{Z} & = & \alpha(\tau)(\bm{x}\bm{x}^t-\bm{Z}).\label{2nd_monent_update}
\end{eqnarray}

First-order moments are given by
\begin{equation}
  E(\bm{x}) = \bm{\mu},
\end{equation}
and second-order moments are given by
\begin{eqnarray}
 E(\bm{x}\bm{x}^t) & = & \bm{\mu}\bm{\mu}^t + E[(\bm{x}-\bm{\mu})(\bm{x}-\bm{\mu})^t],\nonumber \\
 & = & \bm{\mu}\bm{\mu}^t + \bm{\Sigma}.
\end{eqnarray}

The covariance matrix is estimated by $\bm{\Sigma} = \bm{Z}-\bm{z}\bm{z^t}$ using the method of moments; thus, the update rule is as follows:
\begin{eqnarray*}
 \bm{\Sigma} + \Delta \bm{\Sigma} & = & (\bm{Z}+\Delta\bm{Z}) - (\bm{z}+\Delta\bm{z})(\bm{z}+\Delta\bm{z})^t, \\
 & = & (\bm{Z} - \bm{z}\bm{z}^t) + \Delta\bm{Z} - \Delta\bm{z}\Delta\bm{z}^t - \bm{z}\Delta\bm{z}^t - \Delta\bm{z}\bm{z}^t.
\end{eqnarray*}
We then obtain
\begin{equation}
 \Delta\bm{\Sigma} = \Delta\bm{Z}- \Delta\bm{z}\Delta\bm{z}^t - \bm{z}\Delta\bm{z}^t - \Delta\bm{z}\bm{z}^t.\label{145430_24Sep21}
\end{equation}

(\ref{1st_moment_update}) and (\ref{2nd_monent_update}) yield
\begin{eqnarray}
 \Delta\bm{z}\Delta\bm{z}^t & = & \alpha^2 (\bm{x}-\bm{z})(\bm{x}-\bm{z})^t,\label{145350_24Sep21}\\
 \bm{z}\Delta\bm{z}^t & = & \alpha(\bm{z}\bm{x}^t - \bm{z}\bm{z}^t),\label{152240_24Sep21}\\
 \Delta\bm{z}\bm{z}^t & = & \alpha(\bm{x}\bm{z}^t - \bm{z}\bm{z}^t).\label{145407_24Sep21}
\end{eqnarray}

Substituting (\ref{2nd_monent_update}), (\ref{145350_24Sep21}), (\ref{152240_24Sep21}), and (\ref{145407_24Sep21}) for (\ref{145430_24Sep21}), we obtain the following:

\begin{eqnarray}
 \Delta \bm{\Sigma} & = & \alpha (\bm{x}\bm{x}^t - \bm{Z}) - \alpha^2 (\bm{x}-\bm{z})(\bm{x}-\bm{z})^t -  \alpha(\bm{z}\bm{x}^t - \bm{z}\bm{z}^t) - \alpha(\bm{x}\bm{z}^t - \bm{z}\bm{z}^t) \nonumber \\
 & = & \alpha(1-\alpha)(\bm{x}-\bm{z})(\bm{x}-\bm{z})^t - \alpha(\bm{Z}-\bm{z}\bm{z}^t) \nonumber \\
 & = & \alpha [(1-\alpha)(\bm{x}-\bm{z})(\bm{x}-\bm{z})^t - \bm{\Sigma}]\label{151828_24Sep21}
\end{eqnarray}

Therefore, by replacing $\bm{z}$ with $\bm{\mu}$ in (\ref{1st_moment_update}) and (\ref{151828_24Sep21}), we obtain (\ref{171320_24Jan17}) and (\ref{151915_24Sep21}), respectively.

\section{Adjusted Rand index}
\label{ARI}
Let $\bm{S} = \{1,\,2,\,\ldots ,\,n\}$ be the set of indices of $n$.
samples. Let $\mathcal{U}=\{\bm{u}_u\}_{u=1}^U$ and
$\mathcal{V}=\{\bm{v}_v\}_{v=1}^V$ be two different partitions of
$\bm{S}$, where $\bm{u}_u$ and $\bm{v}_v$ are subsets of $\bm{S}$ and
satisfy the following: $\bm{S}=\bigcup _{u=1}^U \bm{u}_u = \bigcup _{v=1}^V
\bm{v}_v$. $u\neq u' \Rightarrow \bm{u}_u \cap \bm{u}_{u'}=\emptyset$
and $v\neq v' \Rightarrow \bm{v}_v \cap \bm{v}_{v'}=\emptyset$.

Considering a sample pair $\{i,\,j\}\subseteq \bm{S}$ and the following calculation,
\begin{eqnarray}
 {\rm TP} & = & \#\{\{i,\,j\}\mid \{i,j\}\subseteq \bm{u}_u \wedge
  \{i,j\}\subseteq \bm{v}_v\},\\
 {\rm FP} & = &  \#\{\{i,\,j\}\mid \{i,j\}\not\subseteq \bm{u}_u \wedge
  \{i,j\}\subseteq \bm{v}_v\},\\
 {\rm FN} & = &  \#\{\{i,\,j\}\mid \{i,j\}\subseteq \bm{u}_u \wedge
 \{i,j\}\not\subseteq \bm{v}_v\},\\
 {\rm TN} & = &  \#\{\{i,\,j\}\mid \{i,j\}\not\subseteq \bm{u}_u \wedge
 \{i,j\}\not\subseteq \bm{v}_v\},
\end{eqnarray}
then the Rand index (RI) is given by
\begin{equation}
  {\rm RI} =  \frac{\rm TP+TN}{\rm TP+FP+FN+TN} = \frac{\rm TP+TN}{{n \choose 2}}.
\end{equation}

The ARI is then defined as
\begin{equation}
 {\rm ARI} = \frac{{\rm RI} - E[\,\rm{RI}\,]}{1 -
  E[\,\rm RI]\,},
\end{equation}
where $E[\,{\rm RI}\,]$ is the expected value of the RI when the two partitions
$\mathcal{U}$ and $\mathcal{V}$ are independent, given by
\begin{eqnarray}
  E[\,{\rm RI}\,]& = &
  \left.
   1+2\sum_{u=1}^U {|\bm{u}_u| \choose 2}\sum _{v=1}^V{|\bm{v}_v|
   \choose 2} \middle/  {n \choose 2}^2   \right. \nonumber \\
   & & \left. - \left[\sum_{u=1}^U {|\bm{u}_u| \choose 2} + \sum
    _{v=1}^V{|\bm{v}_v| \choose 2} \right] \middle/ {n \choose 2}
       \right. .
\end{eqnarray}

\section{Normalized mutual information}
\label{NMI}
Let $\bm{S} = \{1,\,2,\,\ldots ,\,n\}$ be the set of indices of $n$
samples. Let $\mathcal{U}=\{\bm{u}_u\}_{u=1}^U$ and
$\mathcal{V}=\{\bm{v}_v\}_{v=1}^V$ be two different partitions of
$\bm{S}$, where $\bm{u}_u$ and $\bm{v}_v$ are subsets of $\bm{S}$ and
satisfy the following: $\bm{S}=\bigcup _{u=1}^U \bm{u}_u = \bigcup _{v=1}^V
\bm{v}_v$. $u\neq u' \Rightarrow \bm{u}_u \cap \bm{u}_{u'}=\emptyset$
and $v\neq v' \Rightarrow \bm{v}_v \cap \bm{v}_{v'}=\emptyset$.

According to information theory, the mutual information between $\mathcal{U}$ and $\mathcal{V}$ is calculated as follows:

\begin{equation}
 I(\mathcal{U}, \mathcal{V}) = \sum_{u=1}^U\sum_{v=1}^V P(u, v) \log \left(\frac{P(u, v)}{P(u)P(v)}\right),
\end{equation}
where $P(u)=|\bm{u}_u|/n$, $P(v)=|\bm{v}_v|/n$, and $P(u, v)=|\bm{u}_u \cap \bm{v}_v|/n$.

The NMI is defined as follows:
\begin{equation}
 {\rm NMI}(\mathcal{U}, \mathcal{V}) = \frac{I(\mathcal{U}, \mathcal{V})}{{\rm max}\{H(\mathcal{U}), H(\mathcal{V})\}},
\end{equation}
where $H(\mathcal{U})$ and $H(\mathcal{V}))$ are the entropies of $\mathcal{U}$ and $\mathcal{V}$, respectively, defined as follows:
\begin{eqnarray}
 H(\mathcal{U}) & = & -\sum_{u=1}^U P(u)\log(P(u))\\
 H(\mathcal{V}) & = & -\sum_{v=1}^V P(v)\log(P(v))
\end{eqnarray}

\bibliographystyle{elsarticle-num}
\bibliography{ref}

\end{document}